\documentclass{article} 

\usepackage{amsmath,amsfonts,bm}

\def\eqref#1{equation~\ref{#1}}

\def\1{\bm{1}}

\def\rmI{{\mathbf{I}}}

\def\vu{{\bm{u}}}
\def\vv{{\bm{v}}}

\def\vx{{\bm{x}}}
\def\vy{{\bm{y}}}
\def\vz{{\bm{z}}}

\DeclareMathAlphabet{\mathsfit}{\encodingdefault}{\sfdefault}{m}{sl}
\SetMathAlphabet{\mathsfit}{bold}{\encodingdefault}{\sfdefault}{bx}{n}

\usepackage{pifont}
\usepackage{hyperref}
\usepackage{url}
\usepackage{graphicx}
\usepackage[utf8]{inputenc} 
\usepackage[T1]{fontenc}    
\usepackage{booktabs}       
\usepackage{mathtools,amssymb}
\usepackage{amsfonts}       
\usepackage{nicefrac}       
\usepackage{microtype}      
\usepackage{pgfplots,pgfplotstable}
\pgfplotsset{compat=1.14}
\usepackage{array,colortbl}
\usepackage{xcolor}
\usepackage[capitalise]{cleveref}
\usepackage{caption}
\usepackage{graphbox}
\usepackage{placeins}
\usepackage{wrapfig}
\usepackage{subcaption}
\usepackage{etoolbox}
\usepackage{bbm}
\usepackage{amsmath,amssymb}
\usepackage{blindtext}
\usepackage{animate}
\usepackage{array}

\hypersetup{
    colorlinks=true,
    linkcolor = blue,
    anchorcolor = blue,
    citecolor = blue,
    filecolor = blue,
    urlcolor = blue
}

\newcommand{\ie}{\textit{i.e. }}
\newcommand{\eg}{\textit{e.g. }}
\newcommand{\model}{INRFlow}
\usepackage{microtype}
\usepackage{graphicx}
\usepackage{booktabs} 

\usepackage[accepted]{icml2025}

\usepackage{amsmath}
\usepackage{amssymb}
\usepackage{mathtools}
\usepackage{amsthm}

\theoremstyle{plain}

\theoremstyle{definition}

\theoremstyle{remark}

\usepackage[textsize=tiny]{todonotes}

\icmltitlerunning{INRFlow: Flow Matching for INRs in Ambient Space}

\begin{document}

\twocolumn[
\icmltitle{INRFlow: Flow Matching for INRs in Ambient Space}



\icmlsetsymbol{equal}{*}

\begin{icmlauthorlist}
\icmlauthor{Yuyang Wang}{apple}
\icmlauthor{Anurag Ranjan}{apple,left}
\icmlauthor{Josh Susskind}{apple}
\icmlauthor{Miguel Angel Bautista}{apple}

\end{icmlauthorlist}

\icmlaffiliation{apple}{Apple, Machine Learning Research}
\icmlaffiliation{left}{Work done while at Apple}

\icmlcorrespondingauthor{Yuyang Wang}{yuyangw@apple.com}
\icmlcorrespondingauthor{Miguel Angel Bautista}{mbautistamartin@apple.com}

\icmlkeywords{Machine Learning, ICML}

\vskip 0.3in
]



\printAffiliationsAndNotice{}  

\begin{abstract}
Flow matching models have emerged as a powerful method for generative modeling on domains like images or videos, and even on irregular or unstructured data like 3D point clouds or even protein structures. These models are commonly trained in two stages: first, a data compressor is trained, and in a subsequent training stage a flow matching generative model is trained in the latent space of the data compressor. This two-stage paradigm sets obstacles for unifying models across data domains, as hand-crafted compressors architectures are used for different data modalities. To this end, we introduce \textbf{\model}, a domain-agnostic approach to learn flow matching transformers directly in ambient space. Drawing inspiration from INRs, we introduce a conditionally independent point-wise training objective that enables {\model} to make predictions continuously in coordinate space. Our empirical results demonstrate that {\model} effectively handles different data modalities such as images, 3D point clouds and protein structure data, achieving strong performance in different domains and outperforming comparable approaches. {\model} is a promising step towards domain-agnostic flow matching generative models that can be trivially adopted in different data domains. 
\end{abstract}


\section{Introduction}

Recent advances in generative modeling have enabled learning complex data distributions by combining both powerful architectures and training objectives. In particular, state-of-the-art approaches for image ~\citep{sd3}, video \citep{emu} or 3D point cloud ~\citep{lion} generation are based on the concept of iteratively transforming data into Gaussian noise. Diffusion models were originally proposed following this idea and pushing the quality of generated samples in many different domains, including images \citep{emu,ldm,preechakul2022diffusion}, 3D point clouds \citep{ddpm_pointcloud}, graphs \citep{molecules} and video \citep{ddpm_video}. More recently, flow matching \citep{lipman2023flow} and stochastic interpolants \citep{sit} have been proposed as generalized formulations of the noising process, moving from stochastic gaussian diffusion processes to general paths connecting a base (\eg Gaussian) and a target (\eg data) distribution. 

\begin{figure*}[!t]
    \centering
    \includegraphics[width=0.9\textwidth]{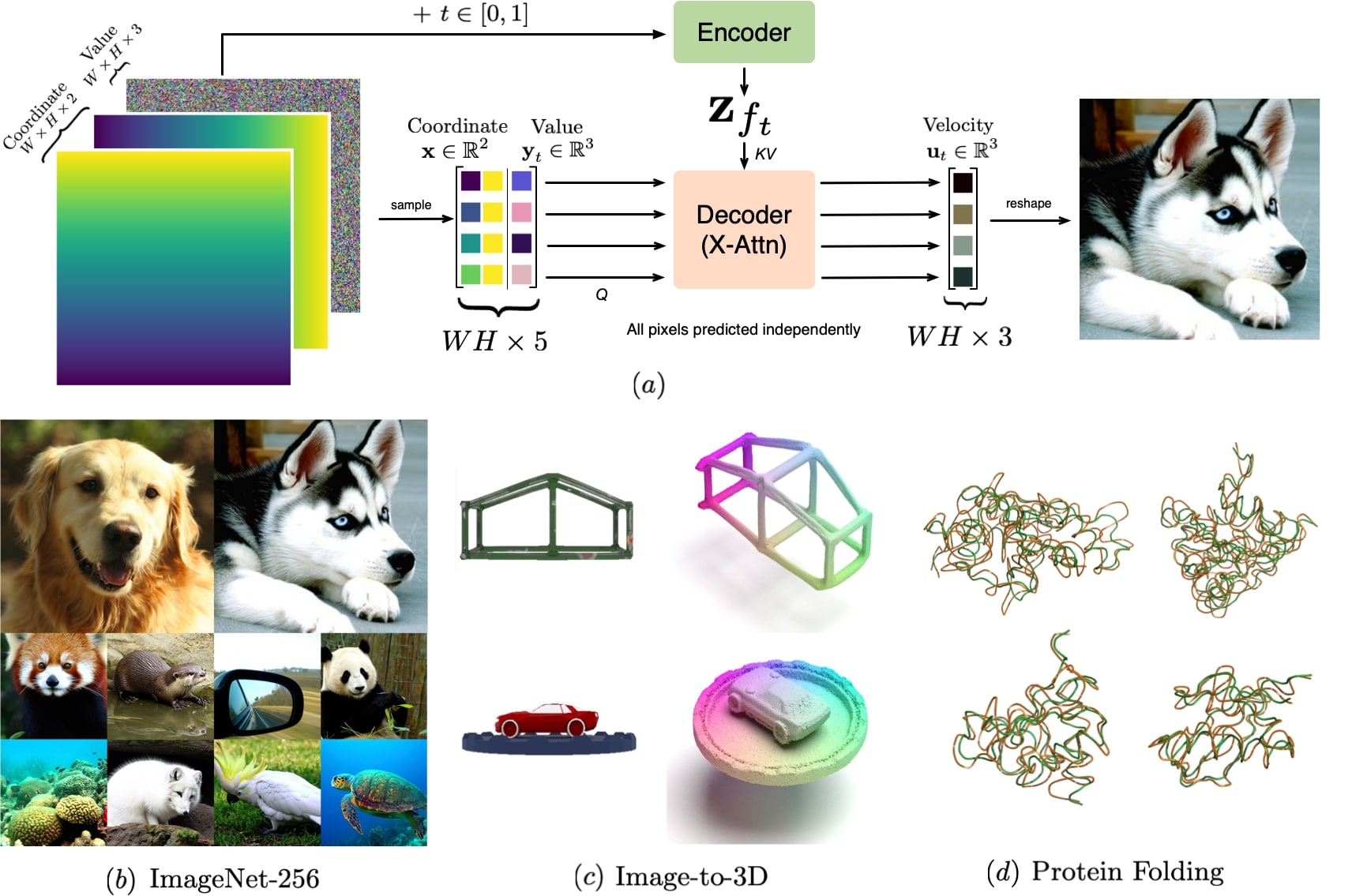}
    \caption{(a) High level overview of {\model} using the image domain as an example. Our model can be interpreted as an encoder-decoder model where the decoder makes predictions independently for each coordinate-value pair given $\vz_{f_t}$. For different data domains, the coordinate and value dimensionality changes, but the model is kept the same. (b) Samples generated by {\model} trained on ImageNet 256$\times$256. (c) Image-to-3D point clouds generated by training {\model} on Objaverse \citep{deitke2023objaverse}. (d) Protein structures generated by {\model} trained on SwissProt \citep{boeckmann2003swiss}. GT protein structures are depicted in green while the generated structures by {\model} are show in orange.}
    \label{fig:intro}
\end{figure*}

In practice, these iterative refinement approaches are commonly applied in a \textit{latent space} obtained from a pre-trained compressor model. Thus, the training recipe consists of two independent training stages: \textit{compressor} (auto-encoder) training and subsequent \textit{generative modeling} in latent space. General purpose transformer architectures have been used in the generative modeling step in latent space \citep{peebles2023scalable, sit, sd3}. However, the first stage compressor uses architectures that are specific to the data domain, requiring hand-crafted inductive biases (\ie ConvNets for image data \citep{ldm}, PointNet for point clouds \citep{lion}, Evoformer for protein structures \cite{alphafold2}).

We see this as one of the core issues preventing the ML community to develop truly domain-agnostic generative models that can be applied to different data domains in a trivial manner. Our goal in this paper is to provide a powerful single training stage approach that is domain-agnostic and simple to implement in practice, thus dispensing with the complexities of two-stage training recipes and enabling modeling of different data modalities directly in ambient (\ie data) space.


It is worth noting that training diffusion or flow matching models in ambient space is indeed possible when using domain specific architecture designs and training recipes. In the image domain, approaches have exploited its dense nature and applied cascaded U-Nets \cite{ho2021cascaded, ho2022cascaded}, joint training of U-Nets at multiple resolutions \cite{gu2023matryoshka}, multi-scale losses \citep{hoogeboom2023simple} or U-Net transformer hybrids architectures \citep{hdit}, obtaining strong results. However, developing strong domain-agnostic models, using general purposes architectures that can be applied across different data domains remains an important open problem.


In this paper, we answer a three part question: \textit{Can we learn flow matching models in \textbf{ambient space}, in a \textbf{single training stage} and using a \textbf{domain agnostic} architecture?} Our goal is to unify different data domains under the same training recipe. To achieve this, we introduce INRFlow {(\model)}, see Fig. \ref{fig:intro}(a). {\model} makes progress towards the goal of unifying flow matching generative modeling across data domains. Drawing inspiration from INRs, we formulate a conditionally independent point-wise training objective that enables training directly in ambient space and can be densely (\eg continuously) evaluated during inference. In the image domain, this means that {\model} models the probability of a pixel value given its coordinate (\ie \textit{the probabilistic extension of an INR}), allowing to generate images at different resolution than the one used during training (see Fig. \ref{fig:resolution_agnostic}(a)). We show generated samples from {\model} trained on ImageNet-256 in Fig. \ref{fig:intro}(b), image-to-3D on Objaverse in Fig. \ref{fig:intro}(c) and protein structures on SwissProt Fig. \ref{fig:intro}(d) (see additional samples in Fig.~\ref{fig:imagenet_samples1}, \ref{fig:app_objaverse}, \ref{fig:app_folding}). Our contributions are summarized as follows:

\begin{itemize}
    \item We propose {\model}, a flow matching generative transformer that works on ambient space to enable single stage generative modeling on different data domains. 
    \item Our results show that {\model}, though domain-agnostic, achieves competitive performance on image and 3D point cloud generation compared with strong domain-specific baselines. 
    \item Our point-wise training objective allows for efficient training via sub-sampling dense domains like images while also enabling resolution changes at inference time.
\end{itemize}

\section{Related Work}

Diffusion models have been the major catalyzer of progress in generative modeling, these approaches learn to reverse a forward process that gradually adds Gaussian noise to corrupt data samples \citep{ddpm}. Diffusion models are notable for their simple and robust training objective. Extensive research has explored various formulations of the forward and backward processes \citep{ddim, heat_ddpm, cold_diffusion}, particularly in the image domain. In addition, different denoising networks have been proposed for different data domains like images \citep{improved_ddpm}, videos \citep{ddpm_video}, and geometric data \citep{ddpm_pointcloud}.  More recently, flow matching \citep{liu2022flow, lipman2023flow} and stochastic interpolants \citep{sit} have emerged as flexible formulations that generalized Gaussian diffusion paths, allowing to define different paths to connect a base and a target distribution. These types of models have shown incredible results in the image domain \citep{sit, sd3} when coupled with transformer architectures \citep{transformers} to model distributions in latent space learnt by data compressors \citep{peebles2023scalable, sit, ldm, lion, maskdit, mdt}. Note that these data compressors use domain specific architectures with hand-crafted inductive biases.

In an attempt to unify generative modeling across various data domains, continuous data representations (also referred to as \textit{implicit neural representation}, \textit{neural fields} or \textit{neural operators}) have shown potential in different approaches: From Data to Functa (Functa) \citep{functa}, Generative Manifold Learning (GEM) \citep{gem}, and Generative Adversarial Stochastic Process (GASP) \citep{gasp} have studied the problem of generating continuous representations of data. More recently Infinite Diffusion \citep{infdiff} and PolyINR \citep{polyinr} have shown great results in the image domain by modeling images as continuous functions. However, both of these approaches make strong assumptions about image data. In particular, \citep{infdiff} interpolates sparse pixels to an euclidean grid to then process it with a U-Net. On the other hand, \citep{polyinr} uses a patching and 2D convolution in the discriminator. Our approach also relates to DPF \cite{dpf}, a diffusion model that acts on function coordinates and can be applied in different data domains on a grid at low resolutions (\ie 64$\times$64). Our approach  is able to deal with higher resolution functions (\eg 256x256 vs. 64x64 resolution images) on large scale datasets like ImageNet, while also tackling unstructured data domains that do not live on an Euclidean grid (\eg like 3D point clouds and protein structures).

\section{Method}
\label{sect:method}

\subsection{Data as Continuous Coordinate $\rightarrow$ Value Maps}

We interpret our empirical data distribution $q$ to be composed of maps $f \sim q(f)$. These maps take \textit{coordinates} $\vx$ as input  to \textit{values} $\vy$ as output. For images, maps are defined from 2D pixel coordinates $ \vx \in \mathbb{R}^2$ to corresponding RGB values $\vy \in \mathbb{R}^3$, thus $f: \mathbb{R}^2 \rightarrow \mathbb{R}^3$, where each image is a different map. For 3D point clouds, $f$ can be interpreted as a deformation that maps coordinates from a fixed base configuration in 3D space to a deformation value also in 3D space, $f: \mathbb{R}^3 \rightarrow \mathbb{R}^3$, as in the image case, each 3D point cloud corresponds to a different deformation map $f$. For ease of notation, we define coordinates $\vx$ and values $\vy$ of any given map $f$ as $\vx_f$ and $\vy_f$, respectively. Fig. \ref{fig:intro}(a) shows an example of such maps in the image domain.

In practice, analytical forms for these maps $f$ are unknown. In addition, different from previous approaches \citep{functa, gem}, we do not fit separate INRs to each data sample via reconstruction, since that would involve a separate training stage fitting an MLP for each map \citep{functa, spatialfuncta, gem}. As a result, we assume we are only given sets of corresponding \textit{coordinate} and \textit{value} pairs resulting from observing these maps at a particular sampling rate (\eg at a particular resolution in the image case). In the following, we develop an end-to-end approach that can directly take these coordinate-value sets as training data and train a model that extends INRs to the probabilistic setting.

\subsection{Flow Matching and Stochastic Interpolants}
We consider generative models that learn to reverse a time-dependent forward process that turns data samples (\ie maps $f$ in our case) $f \sim q(f)$ into noise $\epsilon \sim \mathcal{N}(0, \rmI)$. 

\begin{equation}
    \label{eq:time_depedent_process}
    f_t = \alpha_t f + \sigma_t \epsilon
\end{equation}

Both flow matching \citep{lipman2023flow} and stochastic interpolant \citep{sit} formulations build this forward process in Eq. \ref{eq:time_depedent_process} so that it interpolates exactly between data samples $f$ at time $t=0$ and $\epsilon$ at time $t=1$, with $t \in [0, 1]$. In particular, $p_1(f) \sim \mathcal{N}(0, \rmI)$ and $p_0(f) \approx q(f)$. In this case, the marginal probability distribution $p_t(f)$ of $f$ is equivalent to the distribution of the probability flow ODE with the following velocity field \citep{sit}: 

\begin{equation}
    \label{eq:prob_flow}
    d_t{f_t} = \vu_t(f_t) d_t
\end{equation}

where the velocity field is given by the following conditional expectation, 

\begin{eqnarray}
    \vu_t(f) = \mathbb{E}[d_t{f_t} | f_t = f]=  \nonumber
    \\ = d_t \alpha_t \mathbb{E}[f_0 | f_t = f ] + d_t \sigma_t \mathbb{E}[\epsilon | f_t = f].
\end{eqnarray}

Under this formulation, samples $f_0 \sim p_0(f)$ are generated by solving the probability flow ODE in Eq. \ref{eq:prob_flow} backwards in time (\eg. flowing from $t=1$ to $t=0$), where $p_0(f) \approx q(f)$. Note that both the flow matching \citep{lipman2023flow} and stochastic interpolant \citep{sit} formulations decouple the time-dependent process formulation from the specific choice of parameters $\alpha_t$ and $\sigma_t$, allowing for more flexibility. Throughout the presentation of our method we will assume a rectified flow \citep{liu2022flow,lipman2023flow} or linear interpolant path \citep{sit} between noise and data, which define a straight path to connect data and noise: $f_t =(1 - t) f_0 + t \epsilon$. Note that our framework for learning flow matching models for coordinate-value sets can be used with any path definition. Compared with diffusion models~\citep{ddpm}, linear flow matching objectives result in better training stability and more modeling flexibility~\citep{sit,sd3} which we observed in our early experiments.

\subsection{INRFlow}

We now turn to the task of formulating a flow matching training objective for data distributions of maps $f$. We recall that in practice we do not have access to an analytical or parametric form for these maps $f$ (\eg we do not requite to pretrain INRs for each training sample), and we are only given sets of corresponding \textit{coordinate} $\vx_f$ and \textit{value} $\vy_f$ pairs resulting from observing the mapping at a particular rate. As a result, we need to formulate a training objective that can take these sets of coordinate-value as training data.

In order to achieve this, we first observe that the target velocity field $\vu_t(f_t) d_t$ can be decomposed across both the domain and co-domain of $f_t$, resulting in  a \textit{point-wise velocity field} $\vu_t(\vx_{f_t}, \vy_{f_t}) d_t$, defined for corresponding coordinate and value pairs of $f_t$. As an illustrative example in the image domain, this means that the \textit{target velocity field can be independently evaluated} for any pixel coordinate $\vx_{f_t}$ with corresponding value $\vy_{f_t}$, so that $\vu_t(\vx_{f_t}, \vy_{f_t}) \in \mathbb{R}^3$. Note that one can always decompose target velocity fields in this way since the time-dependent forward process in Eq. \ref{eq:time_depedent_process} aggregates data and noise \textit{independently} (\eg point-wise) across the domain of $f$. Again, using the image domain as an example, the time-dependent forward process of a pixel at coordinate $\vx_f$ is not dependent on other pixel positions or values.

Our goal now is to formulate a training objective to match this point-wise independent velocity field. We want our neural network $\vv_\theta$  parametrizing the velocity field to be able to independently predict a velocity for any given coordinate and value pair $\vx_{f_t}$ and $\vy_{f_t}$. However, this point-wise independent prediction is futile without access to additional contextual conditioning information about the underlying function $f_t$ at time $t$. This is because even if the forward process is point-wise independent, real data exhibits strong dependencies across the domain $f$ that need to be captured by the model. For example, in the image domain, pixels are not independent from each other and natural images show strong both short and long spatial dependencies across pixels. In order to solve this, we introduce a latent variable $\vz_{f_t}$ that encodes contextual information from a set of given coordinate and value pairs of $f_t$. This contextual latent variable allows us to formulate the learnt velocity field to be \textit{conditionally independent} for coordinate-value pairs given $\vz_{f_t}$. The final point-wise conditionally independent CFM loss, which we denote as CICFM loss is defined as:

\begin{equation}
\mathbb{E}_{t \sim \mathcal{U}[0, 1], f \sim q(f)} || \vv_{\theta}(\vx_{f_t}, \vy_{f_t}, t | \vz_{f_t})  - \vu_t(\vx_f, \vy_f | \epsilon) ||_{2}^{2},
\end{equation}

where the target velocity field $\vu_t(\vx, \vy | \epsilon)$ is defined as a rectified flow \citep{liu2022flow,lipman2023flow,sit}: $\vu_t(\vx_f, \vy_f | \epsilon)  = \frac{\epsilon - \vy_{f_t}}{1-t}$.

One of the core challenges of learning this type of generative models is obtaining a latent variable $\vz_{f_t}$ that effectively captures intricate dependencies across the domain of the function, specially for high resolution stimuli like images. In particular, the architectural design decisions are extremely important to ensure that $\vz_{f_t}$ does not become a bottleneck during training. In the following we review our proposed architecture.  

\subsection{Network Architecture}

\begin{figure*}[t]
    \centering
    \includegraphics[width=0.9\textwidth]{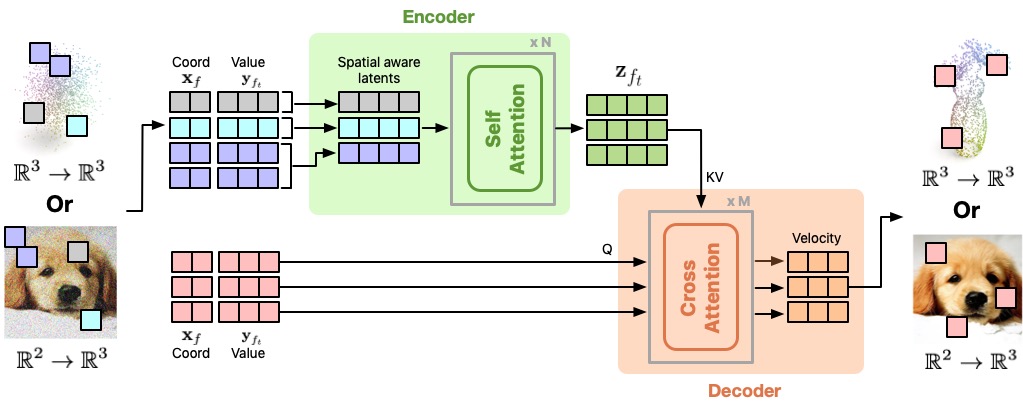}
    \caption{Architecture of our proposed {\model} for different data domains including images and 3D point clouds. \textcolor{black}{Note that models are trained for each data domain separately.} Each spatial aware latent takes in a subset of neighboring context coordinate-value sets in coordinate space. The latents are then updated through self-attention. Decoded coordinate-value pairs cross attend to the updated latents $\vz_{f_t}$ to decode the corresponding velocity.}
    \label{fig:pipeline}
\end{figure*}

We base our model on the general PerceiverIO design \citep{perceiverio},  Fig.~\ref{fig:pipeline} illustrates the architectural pipeline of {\model}. At a high level, our encoder network takes a set of coordinate-value pairs and encodes them to learnable latents through cross-attention. These latents are then updated through several self-attention blocks to provide the final latents $\vz_{f_t} \in \mathbb{R}^{L \times D}$ . To decode the velocity field for a given coordinate-value pair we perform cross attention to $\vz_{f_t}$, generating the final point-wise prediction for the velocity field $\vv_{\theta}(\vx_{f_t}, \vy_{f_t}, t | \vz_{f_t})$.

The encoder of a vanilla PerceiverIO relies solely on cross-attention to the latents \textcolor{black}{$z_{f_t} \in \mathbb{R}^{L \times D}$} to learn spatial connectivity patterns between input and output elements, which we found to introduce a strong bottleneck during training. To ameliorate this, we make a key modifications to boost the performance. Firstly, our encoder utilizes spatial aware latents where each latent is assigned a ``pseudo'' coordinate. Coordinate-value pairs are assigned to latents based on their distances on coordinate space. During encoding, coordinate-value pairs interact with their assigned latents through cross-attention, this means that each of the $L$ latents only attends to a set of neighboring coordinate-value pairs. Latent vectors are then updated using several self-attention blocks. These changes in the encoder allow the model to effectively utilize spatial information while also saving compute when encoding large coordinate-value sets on ambient space.

\section{Experiments}

We evaluate {\model} on two challenging problems: image generation  (FFHQ-256 \citep{ffhq}, LSUN-Church-256 \citep{yu2015lsun}, ImageNet-128/256 \citep{russakovsky2015imagenet}), image-to-3D point cloud generation (Objaverse \citep{deitke2023objaverse}) and protein folding (SwissProt \citep{boeckmann2003swiss}). Note that we use the same training recipe for all tasks, adapted for changes in coordinate-value pair dimensions in different domains. See App. \ref{app:config} for more implementation details and training settings.

\subsection{Image Generation}

Given that {\model} is an probabilistic extension of INRs we compare it with other generative models of the same type, namely approaches that operate in continuous function spaces. Tab.~\ref{tab:dataset-fids} shows a comparison of different image domain specific, as well as, function space models (\eg approaches that model infinite-dimensional signals). {\model} surpasses other generative models in function space on both FFHQ \citep{ffhq} and LSUN-Church \citep{yu2015lsun} at resolution $256 \times 256$. Compared with generative models designed specifically for images, {\model} also achieves comparable or better performance. When scaling up the model size, {\model}-L demonstrates better performance than all the baselines on FFHQ-256 and Church-256, indicating that {\model} can benefit from increasing model sizes.

\begin{table}[t]
\scriptsize
    \vspace{-0.5em}
    \centering
    \begin{tabular}{lccc}
        \toprule
        Model   & \textbf{FFHQ-256} & \textbf{Church-256} \\
        \midrule
        \textit{Domain specific models} & \\ 
        CIPS \citep{anokhin2021image} & 5.29 & 10.80 \\
        StyleSwin \citep{zhang2022styleswin}  & 3.25 & 8.28 \\
        UT \citep{bond2022unleashing}      & 3.05 & 5.52 \\
        StyleGAN2 \citep{karras2020analyzing}  & 2.35 & 6.21 \\ 
        \midrule
        \textit{Function space models} & \\ 
        GEM \citep{du2021learning}  & 35.62 & 87.57  \\ 
        GASP \citep{dupont2022generative}    & 24.37 & 37.46 \\ 
        $\infty$-Diff \citep{infdiff}   & 3.87 & 10.36 \\
        \textbf{\model}-B (ours) & 2.46 & 7.11 \\
        \textbf{\model}-L (ours) & \textbf{2.18} & \textbf{5.51} \\
        \bottomrule
    \end{tabular}
    \vspace{-0.1em}
    \caption{$\text{FID}_{\text{CLIP}}$ \citep{kynkaanniemi2023role} results for state-of-the-art function space approaches.}
    \label{tab:dataset-fids}
    \vspace{-0.6em}
\end{table}

We also evaluate the performance of {\model} on large scale settings previously untapped for domain agnostic approaches, training {\model} on ImageNet at both 128$\times$128 and 256$\times$256 resolutions. On ImageNet-128, shown in Tab. \ref{tab:FID-imagenet128}, {\model} achieves an FID of $2.73$, which is a a competitive performance in comparison to diffusion or flow-based generative baselines including ADM~\citep{dhariwal2021diffusion}, CDM~\citep{ho2021cascaded}, and RIN~\citep{jabri2023scalable} which use domain-specific architectures for image generation. Besides, comparing to PolyINR~\citep{polyinr} which also operates on function space, {\model} achieves competitive FID, while obtaining better IS, precision and recall. In addition, we report results of {\model} for ImageNet-256 on Tab. \ref{tab:FID-imagenet256}. We observed that {\model} is slightly outperformed by latent space models like DiT \citep{peebles2023scalable} and SiT \citep{sit}. We highlight that these baselines rely on a pre-trained VAE compressor that was trained on datasets (\ie 9.29M images) that are much larger than ImageNet, while {\model} was trained only with ImageNet data. In addition, {\model} achieves better performance than many of the baselines trained only with ImageNet data including  ADM \citep{dhariwal2021diffusion}, CDM \citep{ho2021cascaded} and Simple Diffusion (U-Net) \citep{hoogeboom2023simple} which all use CNN-based architectures specific for image generation. Note that this is consistent with the results show in Tab. \ref{tab:dataset-fids}, where {\model} outperforms all function space approaches. 

When comparing with approaches using transformer architectures we find that {\model} obtains performance comparable to RIN \citep{jabri2022scalable} and HDiT \citep{hdit}, with slightly worse FID and slightly better IS. However, {\model} is a domain-agnostic architecture that can be trivially applied to different data domains like 3D point clouds or protein structure data (see Sect. \ref{sect:objaverse_exp} and Sect. \ref{sect:protein_folding}). For completeness, we also include billion scale U-Net transformer hybrid models, Simple Diffusion (U-ViT 2B) and VDM++ (U-ViT 2B).  We highlight that the simplicity of implementing and training {\model} models in practice, and the trivial extension to different data domains are strong arguments favoring {\model}. Finally, comparing with \eg PolyINR~\citep{polyinr} which is also a function space generative model we also find comparable performance, with slight worse FID but better Precision and Recall. It is worth noting that \citep{polyinr} applies a pre-trained DeiT model as the discriminator \citep{polyinr}. Whereas our {\model} makes no such assumption about the function or pre-trained models, enabling to trivially apply {\model} to other domains like 3D point clouds (see Sect. \ref{sect:objaverse_exp}) or protein folding.

\begin{table}[t]
\scriptsize
\centering
\begin{tabular}{lcccc}
    \toprule
    \multicolumn{5}{l}{\textbf{Class-Conditional ImageNet 128x128}} \\
    \toprule
    Model  & FID$\downarrow$  & IS$\uparrow$ & Prec$\uparrow$ & Rec$\uparrow$ \\
    \midrule
    \textit{Adversarial models} & & & & \\
    BigGAN-deep~\citep{brock2019large} & 6.02  & 145.8 & \textbf{0.86} & 0.35 \\
    PolyINR~\citep{polyinr} & \textbf{2.08 } & 179.0 & 0.70 & 0.45 \\
    \midrule
    \textit{Diffusion models} & & & & \\
    CDM (w/ cfg)~\citep{ho2021cascaded} & 3.52 & 128.0 & - & - \\
    ADM (w/ cfg)~\citep{dhariwal2021diffusion} & 2.97  & 141.3 & 0.78 & \textbf{0.59 }\\
    RIN~\citep{jabri2023scalable} & 2.75  & 144.0 & - & - \\
    \midrule
    \textbf{\model}-XL (ours) (cfg=1.5)  & 2.73  & \textbf{187.6} & 0.80 & 0.58 \\
    \bottomrule
    \end{tabular}
\caption{Benchmarking class-conditional image generation on ImageNet 128x128.}
\label{tab:FID-imagenet128}
\end{table}

\begin{table*}[t]
\small
\begin{center}
\resizebox{0.95\textwidth}{!}{
\begin{tabular}{lcccccccc}
    \toprule
    \multicolumn{5}{l}{\textbf{Class-Conditional ImageNet 256x256}} \\
    \midrule
    Model  & \textcolor{black}{Agnostic} & \textcolor{black}{\# Tr. Samples} & \textcolor{black}{\# Params} & bs $\times$ it. & FID$\downarrow$  & IS$\uparrow$ & Precision$\uparrow$ & Recall$\uparrow$ \\
    \midrule
    \textit{Adversarial models} &  & & & \\
    BigGAN-deep~\citep{brock2019large} & \textcolor{black}{\ding{55}} & \textcolor{black}{1.28M} & \textcolor{black}{-} &-  &  6.95 & 171.4 & \textbf{0.87} & 0.28 \\
    PolyINR~\citep{polyinr} & \textcolor{black}{\ding{55}} & \textcolor{black}{1.28M} & \textcolor{black}{-} &-  & 2.86  &  241.4 & 0.71 & 0.39\\
    \midrule
    \textit{Latent space with pretrained VAE} & & & & \\
    DiT-XL (cfg=1.5)~\citep{peebles2023scalable} & \textcolor{black}{\ding{55}} & \textcolor{black}{\textbf{9.23M}} & \textcolor{black}{675M} &-  & 2.27  & \textbf{278.2} & 0.83 & 0.57 \\
    SiT-XL (cfg=1.5, SDE)~\citep{sit} & \textcolor{black}{\ding{55}}  & \textcolor{black}{\textbf{9.23M}}& \textcolor{black}{675M} & \textbf{1.8B}  & \textbf{2.06} &  270.2 & 0.82 & 0.59\\
    \midrule
    \textit{Ambient space} & & & & \\
    ADM~\citep{dhariwal2021diffusion} & \textcolor{black}{\ding{55}} & \textcolor{black}{1.28M} & \textcolor{black}{554M} & 507M  & 10.94 & 100.9 & 0.69 & \textbf{0.63} \\
    CDM~\citep{ho2021cascaded} & \textcolor{black}{\ding{55}} & \textcolor{black}{1.28M} & \textcolor{black}{-} &-  & 4.88 & 158.7 & - & - \\
    Simple Diff. (U-Net)~\citep{hoogeboom2023simple} & \textcolor{black}{\ding{55}} & \textcolor{black}{1.28M} & -  &\textcolor{black}{-} & 3.76 & 171.6 & - & -\\
    RIN~\citep{jabri2023scalable} & \textcolor{black}{\ding{55}} & \textcolor{black}{1.28M} & \textcolor{black}{410M} & 614M  & 3.42  & 182.0 & - & - \\
    HDiT (cfg=1.3)~\citep{hdit} & \textcolor{black}{\ding{55}} & \textcolor{black}{1.28M} & \textcolor{black}{557M} & 742M  & 3.21  & 220.6 & - & - \\
    Simple Diff. (U-ViT)~\cite{hoogeboom2023simple} & \textcolor{black}{\ding{55}} & \textcolor{black}{1.28M} & \textcolor{black}{\textbf{2B}} & 1.4B  & 2.77  & 211.8 & - & - \\
    VDM++ (U-ViT)~~\citep{kingma2023understanding} & \textcolor{black}{\ding{55}} & \textcolor{black}{1.28M} & \textcolor{black}{\textbf{2B}} & 1.4B  & 2.12  & 267.7 & - & -\\
    \midrule
    \textbf{\model}-XL (ours) (cfg=1.5) & \textcolor{black}{\ding{51}} & \textcolor{black}{1.28M} & \textcolor{black}{733M} & 780M  & 3.74  & 228.8 & 0.82 & 0.52 \\
    \bottomrule
    \end{tabular}
}
\end{center}
\caption{Top performing models for class-conditional image generation on ImageNet 256x256.}
\label{tab:FID-imagenet256}
\end{table*}

\subsection{Image-to-3D}
\label{sect:objaverse_exp}

We also showcase that {\model} can directly integrate conditional information like images. We train an image-to-point-cloud {\model} model on Objaverse~\citep{deitke2023objaverse}, which contains 800k 3D objects of wide variety, to illustrate the capability of {\model} on large-scale 3D generative tasks. In particular, conditional information (i.e., an image) is integrated to our model through cross-attention. We train {\model} with batch size 384 for 500k iterations. During sampling, we use an Euler-Maruyama sampler~\citep{sit} with 500 steps to generate point clouds (see App. \ref{app:objaverse} for more details on architecture and objaverse data generation).

Tab.~\ref{tb:objaverse} shows the performance of {\model} in comparison with recent baselines on Objaverse. We report ULIP-I~\citep{xue2024ulip} and P-FID~\citep{nichol2022point} following CLAY~\citep{zhang2024clay}. PointNet++~\citep{qi2017pointnet, qi2017pointnet++, nichol2022point} is employed to evaluate P-FID. ULIP-I is an analogy to CLIP for text-to-image generation. ULIP-I is measured as the cosine similarity between point-cloud features from ULIP-2 model~\citep{xue2024ulip} and image features from CLIP model~\citep{radford2021clip}. The performance numbers of baseline models are directly borrowed from CLAY~\citep{zhang2024clay}. We calculate the metrics of our {\model} on 10k sampled point clouds. In our case, P-FID is measured on point clouds with 4096 points following Shape-E~\citep{jun2023shap} while ULIP-I is measured on point clouds with 10k points following ULIP-2~\citep{xue2024ulip}. Note that since CLAY~\citep{zhang2024clay} is not open-source, we do not have the access to the exact evaluation setting or the conditional images rendered from Objaverse. But all evaluation settings of {\model} are provided for reproduction purpose. As shown in Tab.~\ref{tb:objaverse}, our {\model} achieves strong performance on image-to-3D generative tasks. Compared to CLAY~\citep{zhang2024clay}, which is a 2-stage latent diffusion model, {\model} demonstrates very strong performance on both ULIP-I and P-FID.

\begin{table}[tb!]
	\label{table:objaverse}
    \small
	\centering
    \begin{tabular}{lccc}
        \toprule
        Model & ULIP-I $\uparrow$ & P-FID $\downarrow$ \\
        \midrule
        Shap-E \citep{jun2023shap} & 0.1307 & - \\
        Michelangelo \citep{zhao2024michelangelo} & 0.1899 & - \\
        CLAY \citep{zhang2024clay} & 0.2066 & 0.9946 \\
        \midrule
        {\model} (ours) & \textbf{0.2976} & \textbf{0.3638} \\
        \bottomrule
    \end{tabular}
\caption{\textcolor{black}{Image-conditioned 3D point cloud generation performance on Objaverse.}}
\label{tb:objaverse}
\end{table}

\begin{figure*}[!t]
    \centering
    \includegraphics[width=\linewidth]{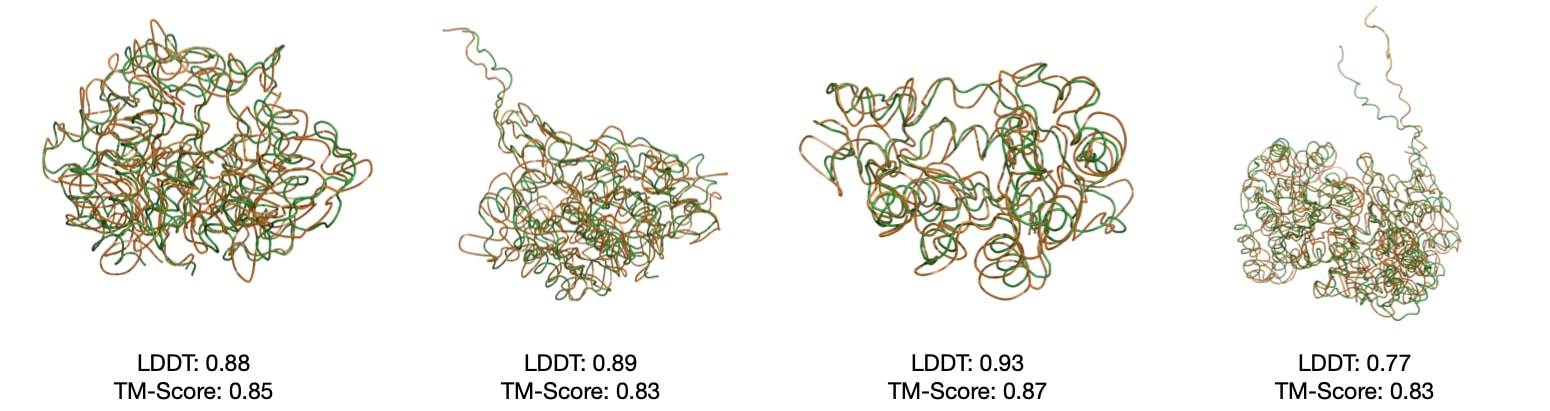}
    \caption{Examples of protein structures predicted by {\model} on SwissProt, together with their LDDT and TM scores. The GT structures are depicted in green while the generated structures are show in orange. {\model} accurately captures the global spatial distribution of protein backbones generating reasonable 3D structures for different protein sequences.}
    \label{fig:protein_folding}
\end{figure*}

Fig.~\ref{fig:resolution_agnostic}(b) show examples of sampled point clouds and corresponding conditional images. As discussed in \S\ref{sec:res_free}, {\model} trained on Objaverse also enjoys the flexibility of resolution agnostic generation. In the App, Fig.~\ref{fig:app_objaverse} we show additional results sampled with more points than what the model was trained on. As shown, {\model} learns to generate 3D objects with rich details that match the conditional images ultimately being able to generate a continuous surface.

\subsection{Protein Folding}
\label{sect:protein_folding}
We now showcase the domain-agnostic prowess of {\model} applying it to the protein folding problem \citep{alphafold2}. From an ML perspective, this problem is a conditional 3D generation problem where we are given the amino-acid sequence (\eg a sequence of discrete symbols from a vocabulary of 20 possible amino-acids) and we need to generate a 3D coordinate for each atom in the protein, where different amino-acids can have different numbers of atoms. In our experiments we use SwissProt set \citep{boeckmann2003swiss} taking the ground truth structures from the AlphaFold Database \citep{afdb}.

We adapt the coordinate and signal in {\model} so that the coordinate becomes a feature representation of each atom in the protein and the signal represents the 3D point for that particular atom. This is similar to recent work on protein folding and conformer generation, which directly predicts atom coordinates as opposed to frames of reference \cite{mcf, alphafold3}. For the spatially-aware latents we aggregate information from all the atoms corresponding to the same amino-acid into a single latent vector (see App. \ref{app:protein_folding} for more details on architecture and training recipe).

We show qualitative results on Fig. \ref{fig:protein_folding} and additional results on Fig. \ref{fig:app_folding}. {\model} accurately captures the global distribution of protein backbones generating reasonable 3D structures for different protein sequences. These initial results are very encouraging and open up exciting design spaces for future work developing protein folding network architectures that can leverage general purpose transformer blocks.

\subsection{Resolution Agnostic Generation}
\label{sec:res_free}

\begin{figure*}[!t]
    \centering
    \includegraphics[width=\linewidth]{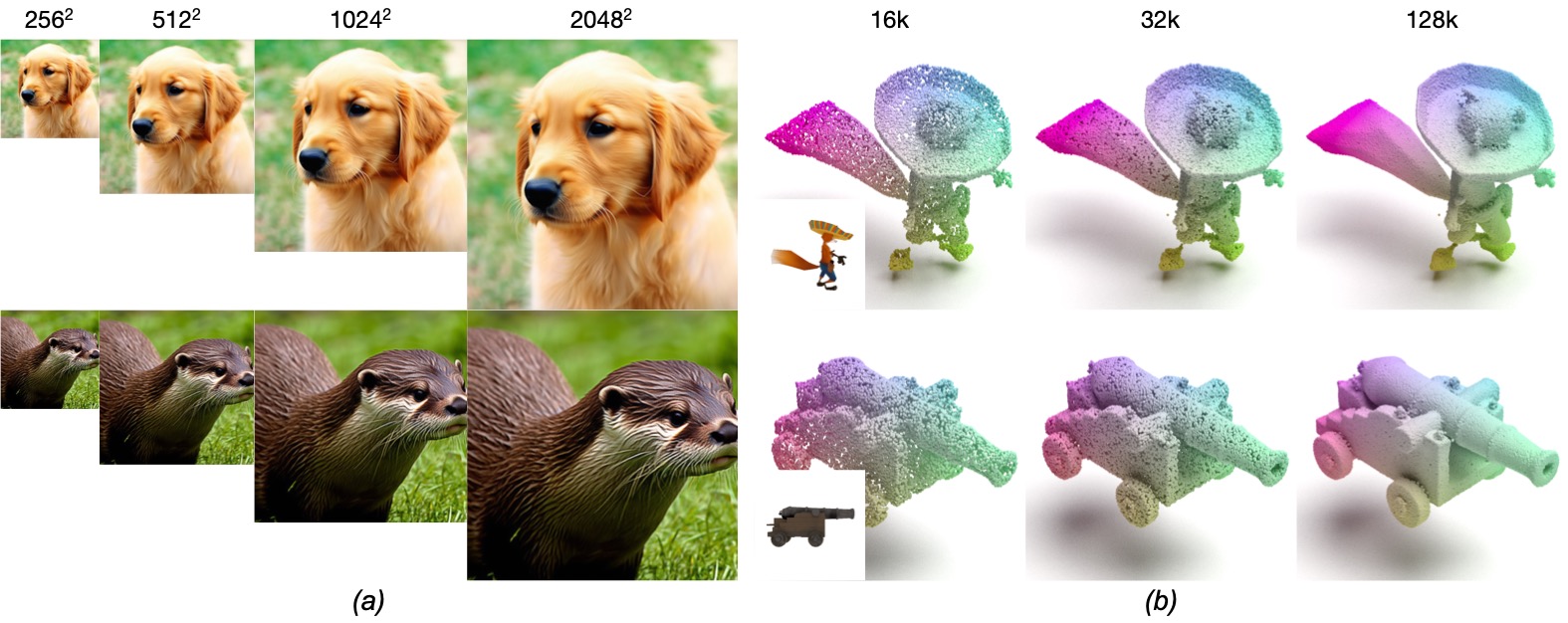}
    \caption{Examples of resolution agnostic generation for {\model} models trained on ImageNet-256 (a),  and Objaverse-16k in (b). To generate samples at higher resolutions than the one in training we fix the initial noise seed and increase the number of coordinate-value pairs evaluated by the model. Even though {\model} was only train with samples at a fixed resolution (256 for ImageNet and 16k for Objaverse), it can still generate realistic samples at higher resolutions. These results show that {\model} is learning a continuous probability density field.}
    \label{fig:resolution_agnostic}
\end{figure*}

An interesting property of {\model} is that it decodes each coordinate-value pair independently, allowing resolution changes of resolution for generation. At inference time the user can define as many coordinate-value pairs as desired where the initial value of each pair at $t=1$ is drawn from a Gaussian distribution. We now quantitatively evaluate the performance of {\model} in this setting. In Tab.~\ref{tab:FID-superres} we compare the FID of different recipes. First, {\model} is trained on FFHQ-256 and bilinear or bicubic interpolation is applied to the generated samples to get images at 512. On the other hand, {\model} can directly generate images at resolution 512 by simply increasing the number of coordinate-value pairs during inference without further tuning. As shown in Tab.~\ref{tab:FID-superres} , {\model} achieves lower FID when compared with other manually designed interpolation methods, showcasing the benefit of developing generative models on ambient space. 

 \begin{table}[h]
  \centering
  \scalebox{0.84}{
  \setlength\tabcolsep{3.5pt}
  \begin{tabular}{lccc}
    \toprule
    & {\model} & Bilinear & Bicubic \\
    \midrule
    FID($\downarrow$) & 23.09 & 35.05 & 24.34 \\
    \bottomrule
  \end{tabular}
  }
  \caption{FID of different super resolution methods to generate images at resolution $512 \times 512$ for {\model} trained on FFHQ-256.}
  \label{tab:FID-superres}
\end{table}

We show examples of resolution agnostic generation for {\model} models trained on ImageNet-256 Fig. \ref{fig:resolution_agnostic}(a) and Objaverse-16k in Fig. \ref{fig:resolution_agnostic}(b). Even though {\model} was only train with samples at a fixed resolution (256 for ImageNet and 16k for Objaverse), it can still generate realistic samples at higher resolutions. For example, Fig. ~\ref{fig:resolution_agnostic}(a) shows samples generated from {\model} trained on ImageNet-256 and sampled at resolutions up to 2k (see additional examples in Fig. \ref{fig:image_upsample_appendix}).  Fig.~\ref{fig:resolution_agnostic}(b) shows point cloud with up to 128k points from {\model} trained on Objaverse with only 16k points points per sample. It is worth noting that in {\model} there's no cascading \citep{ho2022cascaded} or multiple resolution training \citep{gu2023matryoshka}, we simply fix the initial noise seed and increase number of coordinate-value pairs that are evaluated. These results show that {\model} is not trivially overfitting to the coordinate-value pairs in the training but rather learning a continuous probability density field in space from which an infinite number of points could be sampled. Generally speaking, this also provides the potential to efficiently train flow matching generative models without the need to use large amounts of expensive high resolution data, which can be hard to collect in some domains.

\section{Conclusion}

We introduced {\model}, a flow matching generative model for continuous function spaces designed to operate directly in ambient space. Our approach dispenses with the practical complexities of training latent space generative models \cite{gem, functa}, such as the dependence on domain-specific compressors for different data domains or tuning of hyper-parameters of the data compressor (\ie adversarial weight, KL term, etc.). Inspired by deterministic encoding of INRs, we introduced a conditionally independent point-wise training objective that decomposes the target vector field and allows to continuously evaluate the generated samples, similar to INRs, enabling resolution changes at inference time. 

Our results on both image, 3D point cloud and protein folding benchmarks show the strong performance of {\model}, as well as, its trivial adaption across modalities which we believe is the cornerstone of a good generative modeling architecture. In conclusion, {\model} represents a promising direction for flow matching generative models, offering a powerful and domain-agnostic framework. Future work could explore further improvements in training efficiency applying tricks orthogonal to our contribution \citep{microdiffusion} and investigate co-training of multiple data domains to enable multi-modality generation in an end-to-end learning paradigm. 

\newpage

\section*{Impact Statement}
This paper concerns the generative modeling methodology. While we do not see immediate societal implications from our technical contribution, there are potential consequences when it is used in training foundational generative models. 

\bibliography{iclr2024_conference}
\bibliographystyle{iclr2025_conference}


\appendix
\section{Model Configuration and Training Settings}
\label{app:config}

\subsection{Image}
We provide detailed model configurations and training settings of {\model} for image generation in Tab.~\ref{tab:imagenet_model_config}. We develop model sizes small (S), base (B), large (L), and extra large (XL) to approximately match the number of parameters in DiT~\citep{peebles2023scalable}. For image experiments we implement the ``psuedo'' coordinate of latents as 2D grids and coordinate-value pairs are assigned to different latents based on their distances to the latent coordinates. To embed coordinates, we apply standard Fourier positional embedding~\citep{transformers} for ambient space coordinate input in both encoder and decoder. The Fourier positional embedding is also applied to the ``psuedo'' coordinate of latents. On image generation, we found that applying rotary positional embedding (RoPE)~\citep{su2024roformer} slightly improves the performance of {\model}. Therefore, RoPE is employed for largest {\model}-XL model. For all the models we share the following training parameters except the \texttt{training\_steps} across different experiments. On image generation, all models are trained with batch size 256, except for {\model}-XL reported in Tab.~\ref{tab:FID-imagenet128} and Tab.~\ref{tab:FID-imagenet256}, which are trained for 1.7M steps with batch size 512.

\begin{verbatim}
default training config:
    optimizer='AdamW'
    adam_beta1=0.9
    adam_beta2=0.999
    adam_eps=1e-8
    learning_rate=1e-4
    weight_decay=0.0
    gradient_clip_norm=2.0
    ema_decay=0.999
    mixed_precision_training=bf16
\end{verbatim}

In Tab.~\ref{tab:model_config_compare}, we also compare the size of models trained on ImageNet-256, training cost (\ie product of batch size and training iterations), and inference cost (\ie NFE, number of function evaluation). Note that for models that achieve better performance than {\model}, many of them are trained for more iterations. In addition,  at  inference time {\model} applies simple first order Euler sampler with 100 sampling steps, which uses less NFE than many other baselines. 

 \begin{table}[ht]
  \centering
  \scalebox{0.7}{
  \setlength\tabcolsep{4pt}
  \begin{tabular}{lccccccccccc}
    \toprule
    Model & Layers & Hidden size & \#Latents & Heads & Decoder layers & \#Params  \\
    \midrule
    {\model}-S    & 12 & 384  & 1024 & 6  & 1 & 35M \\
    {\model}-B    & 12 & 768  & 1024 & 12 & 1 & 138M \\
    {\model}-L    & 24 & 1024 & 1024 & 16 & 1 & 458M \\
    {\model}-XL   & 28 & 1152 & 1024 & 16 & 2 & 733M \\
    \bottomrule
  \end{tabular}
  }
  \caption{Detailed configurations of {\model} for image generation.}
  \label{tab:imagenet_model_config}
\end{table}

 \begin{table*}[h]
  \centering
  \scalebox{0.84}{
  \setlength\tabcolsep{4pt}
  \begin{tabular}{lcccccc}
    \toprule
    Model & \# Train data&  \# params & bs.$\times$it. & NFE & FID $\downarrow$  & IS $\uparrow$\\
    \midrule
        ADM~\citep{dhariwal2021diffusion} & 1.28M & 554M & 507M & 1000 & 10.94 & 100.9\\
        RIN~\citep{jabri2023scalable} & 1.28M & 410M & 614M & 1000  & 3.42 &  182.0\\
        HDiT~\citep{hdit}&  1.28M  & 557M & 742M & 100 & 3.21 & 220.6 \\
        Simple Diff. (U-ViT 2B)~\citep{hoogeboom2023simple} & 1.28M  & 2B & 1B & - & 2.77 & 211.8 \\
        DiT-XL~\citep{peebles2023scalable} & 9.23M & 675M & 1.8B & 250 & 2.27 & 278.2\\
        VDM++ (U-ViT 2B)~\citep{kingma2023understanding} & 1.28M  & 2B & 1.4B & 512 & 2.12 & 267.7  \\
        SiT-XL~\citep{sit} & 9.23M  & 675M & 1.8B & 500 & 2.06 & 270.2 \\
    \midrule
    {\model}-XL (ours) & 1.28M  & 733M & 870M & 100 & 3.74 & 228.8 \\
    \bottomrule
  \end{tabular}
  }
  \caption{Comparison of {\model} and baselines in \# params and training cost (\ie product of batch size and training iterations). Some numbers are borrowed from \cite{hdit}.}
  \label{tab:model_config_compare}
\end{table*}

\subsection{Image-to-3D}
\label{app:objaverse}

For image-to-3D point cloud generation we trained {\model} on Objaverse~\citep{deitke2023objaverse}, which contains 800k 3D objects of great variety. In particular, conditional information (i.e., an image) is integrated to our model through cross-attention. For each object in Objaverse, we sample point cloud with $16k$ points. To get images for conditioning, each object is rendered with 40 degrees field of view, $448 \times 448$ resolution, at 3.5 units on the opposite sides of $x$ and $z$ axes looking at the origin. We extract features via DINOv2~\citep{oquab2023dinov2} which is concatenated with Plucker ray embedding~\citep{plucker2018analytisch} of each patch in DINOv2 feature. In each block, the learnable latent vector $z_{f_t}$ cross attends to image feature. During training, the image conditioning is dropped randomly with 10\% probability. Therefore, our model can also benefit from popular classifier-free guidance (CFG) to increase the guidance strength. The model is trained
with batch size 384 for 500k iterations. During sampling, we use an Euler-Maruyama sampler with 500 steps to generate point clouds. We train an {\model}-XL size model of 866M parameters similar to the one reported in Tab. \ref{tab:shapenet_model_config}.

One particularity for image-to-3D point cloud generation is that we assign input elements to latents through a hash code, so that neighboring input elements are likely (but not certainly) to be assigned to the same latent token. We found that the improvements of spatial aware latents in 3D to not be as substantial as in the 2D image setting, so we report results with a vanilla PerceiverIO architecture for simplicity. To embed coordinates, we apply standard Fourier positional embedding~\citep{transformers} for ambient space coordinate input in both encoder and decoder.

\subsection{Protein Folding}
\label{app:protein_folding}

In our experiments we use SwissProt set \citep{boeckmann2003swiss} taking the ground truth structures from the AlphaFold Database \citep{afdb}. We select a random set of 10k protein structures to train {\model}. In this setting, the coordinate-value pairs represent atoms in the protein, where the "coordinate" part is a set of features of that particular atom. In particular, we use the atom features shown in Tab. \ref{tab:atom_features} together with the eigenvectors of the graph laplacian for each residue, as previously done in \citep{mcf}. In particular, we found it beneficial to also concatenate to the atomic features the amino-acid embedding of its corresponding amino-acid, which we obtain from a pre-trained ESM model (ESM-650M), a language model for protein sequences masking only on sequence data and not 3D structures. 

We aggregate information into spatially-aware latents by cross-attending all atoms belonging to a particular amino-acid with its particular latent. In practice, since the number of atoms for each aminoacid type is always fixed, one might also simply use a linear layer. Finally, for this task we train a XL size model for 100k iterations with batch size 256.

\begin{table*}[h!]
  \centering
  \small
  \begin{tabular}{>{\color{black}}l >{\color{black}}l >{\color{black}}l}
    \toprule
    Name & Description & Range \\
    \midrule
    \texttt{atomic} & Atom type & one-hot of 35 elements in dataset \\
    \texttt{degree} & Number of bonded neighbors & $\{x:0 \leq x \leq 6, x \in \mathbbm{Z}\}$ \\
    \texttt{charge} & Formal charge of atom & $\{x:-1 \leq x \leq 1, x \in \mathbbm{Z}\}$ \\
    \texttt{valence} & Implicit valence of atom & $\{x:0 \leq x \leq 6, x \in \mathbbm{Z}\}$ \\
    \texttt{hybrization} & Hybrization type & \{sp, sp\textsuperscript{2}, sp\textsuperscript{3}, sp\textsuperscript{3}d, sp\textsuperscript{3}d\textsuperscript{2}, other\} \\
    \texttt{aromatic} & Whether on a aromatic ring & \{True, False\} \\
    \texttt{num\_rings} & number of rings atom is in & $\{x:0 \leq x \leq 3, x \in \mathbbm{Z}\}$ \\
    \texttt{ESM embedding} & Amino-acid embedding & $x \in \mathbb{R}^{1280}$ \\ 
    \bottomrule
  \end{tabular}
  \caption{\textcolor{black}{Atomic features included in INRFlow for protein folding.}}
  \label{tab:atom_features}
\end{table*}


\section{Performance vs Model Size}
\begin{figure}[h]
\centering
    \includegraphics[width=0.47\textwidth]{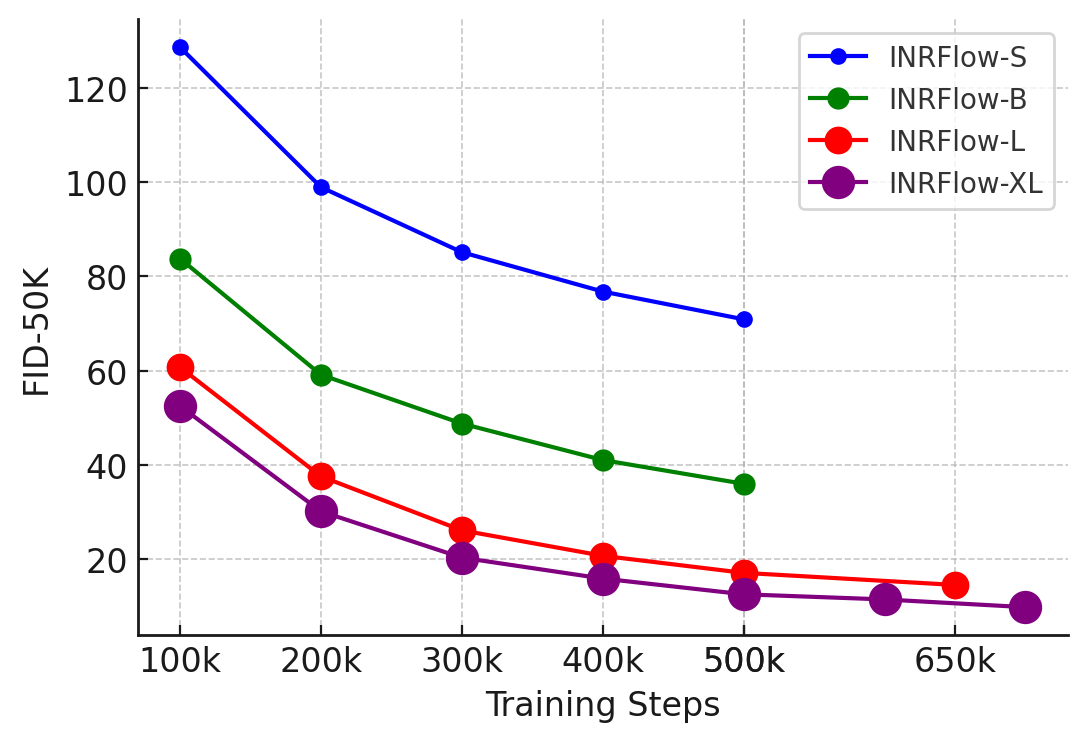} 
    \caption{FID-50K over training iterations with different model sizes, where we see clear benefits of scaling up model sizes.}
    \label{fig:fid_ablation}
\end{figure}

To demonstrate the scalability of {\model} we train models of different sizes including small (S), base (B), large (L), and extra-large (XL) on ImageNet-256. We show the performance of different model sizes using FID-50K in Fig.~\ref{fig:fid_ablation}(a). We observe a clear improving trend when increasing the number of parameters as well as increasing training steps. This demonstrates that scaling the total training Gflops is important to improved generative results as in other ViT-based generative models~\citep{peebles2023scalable, sit}. 

\section{Performance vs Training Compute}
\label{app:gflops}

\begin{figure}[htb!]
\centering
\includegraphics[width=0.45\textwidth]{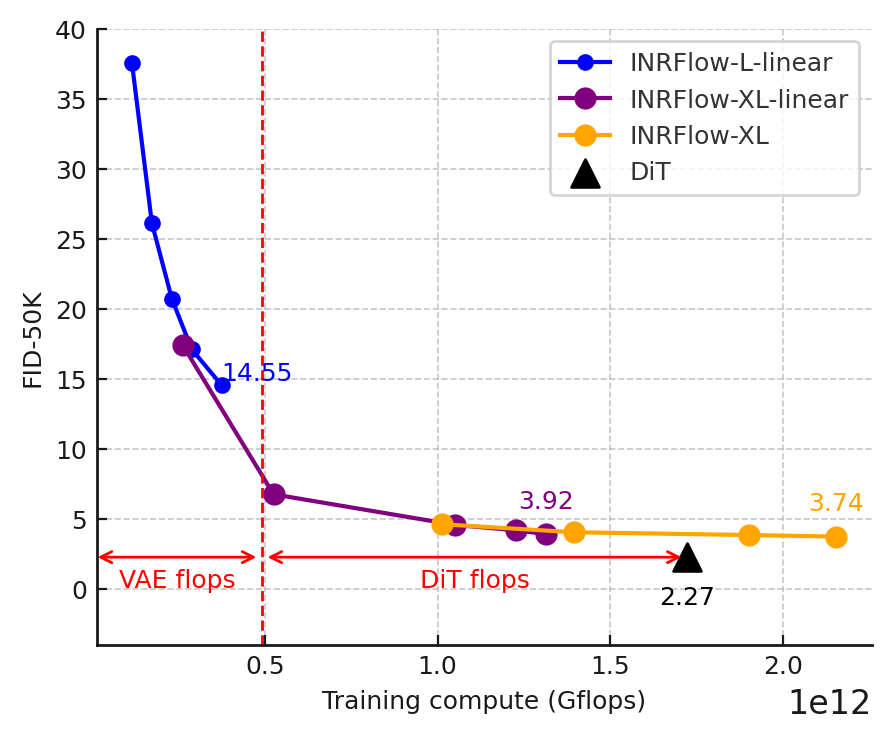}
\caption{Comparing the performance vs total training compute comparison of {\model} and DiT~\citep{peebles2023scalable}. }
\label{fig:gflops_dit_asft}
\end{figure}

We compare the performance vs total training compute of {\model} and DiT \citep{peebles2023scalable} in Gflops. {\model}-linear denotes the variant of {\model} where the cross-attention in the spatial aware encoder is replaced with grouping followed by a linear layer. We found this could be an efficient variant of standard {\model} while still achieving competitive performance. Fig.~\ref{fig:gflops_dit_asft} shows the comparison of the training compute in Gflops vs FID-50K between {\model} and latent diffusion model DiT~\citep{peebles2023scalable} including the tranining compute of the first stage VAE. We estimate the training cost of VAE based the model card listed in HuggingFace\footnote{\url{https://huggingface.co/stabilityai/sd-vae-ft-mse}}. As shown, the training cost of VAE is not negligible and reasonable models with FID $\approx 6.5$ can be trained for the same cost. 

Admittedly, under equivalent training Gflops, {\model} achieves comparable but not as good performance as DiT in terms of FID score (with a difference smaller than $1.65$ FID points). We attribute this gap to the fact that DiT's VAE was trained on a dataset much larger than ImageNet, using a domain-specific architecture (\eg a convolutional U-Net). We believe that the simplicity of implementing and training {\model} models in practice, and the trivial extension to different data domains (as shown in Sect. \ref{sect:shapenet_exp}) are strong arguments to counter an FID difference of smaller than $1.65$ points. In addition, applying masking tricks orthogonal to our approach like the ones in \cite{microdiffusion} can help mitigate the training compute difference.

\begin{figure}[htb!]
\centering
\includegraphics[width=0.45\textwidth]{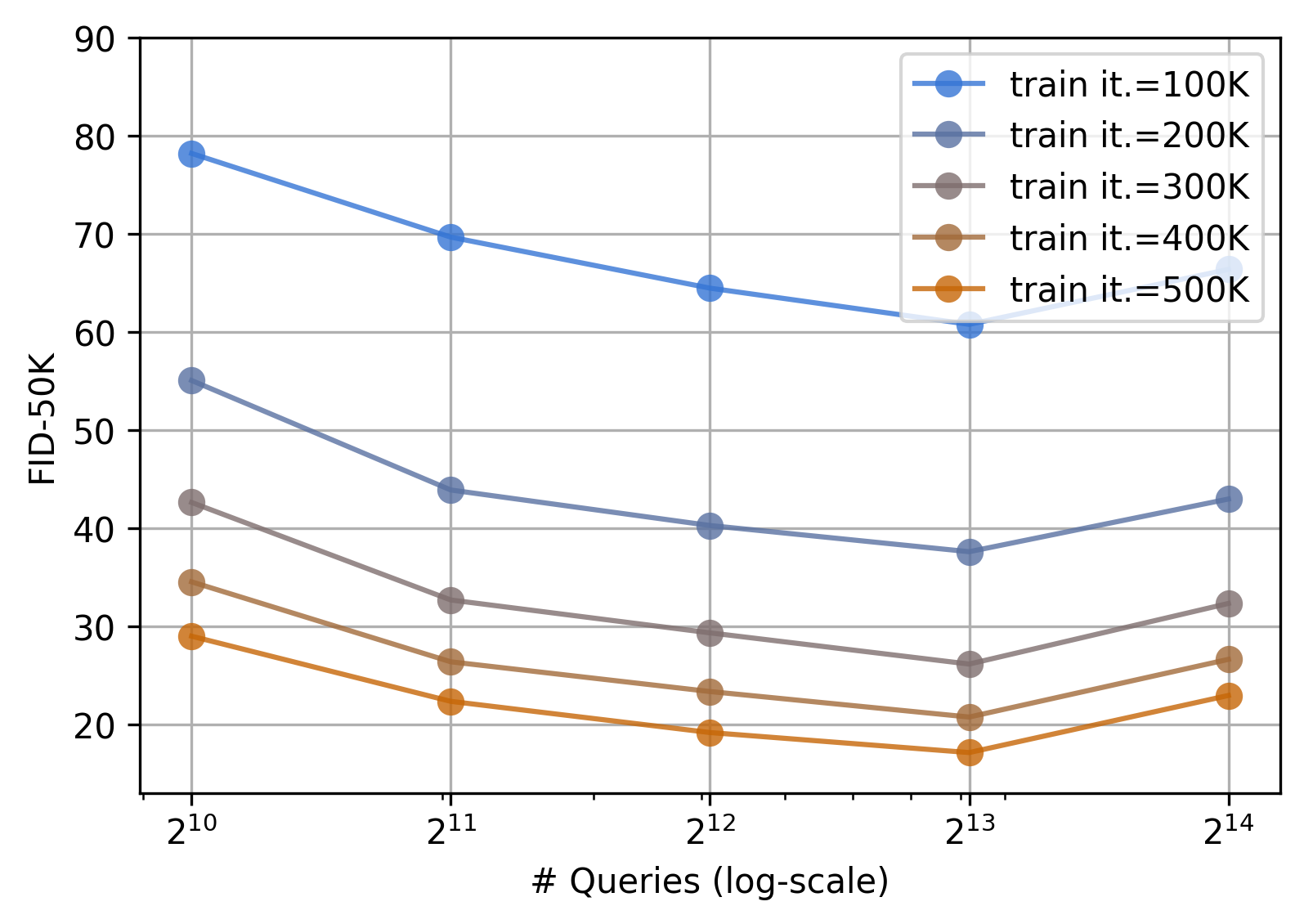}
\caption{ FID-50K over training iterations with different number of decoded coordinate-value pairs during training and the corresponding compute cost for a single forward pass. }
\label{fig:fid_queries}
\end{figure}

In addition, due to the flexibility of cross-attention decoder in {\model}, one can easily conduct random sub-sampling to reduce the number of decoded coordinate-value pairs during training which ca also saves computation. Fig.~\ref{fig:fid_queries} shows how number of decoded coordinate-value pairs affects the model performance as well as Gflops in training. An image of resolution 256$\times$256 contains 65536 pixels in total which is the maximal number of coordinate-value pairs during training. As see in Fig. \ref{fig:fid_ablation}(b), a model decoding 4096 coordinate-value pairs saves more than 20\% Gflops over one decoding 16384. 

\section{Spatial-aware Latent}

 \begin{table}[h]
  \centering
  \scalebox{0.84}{
  \setlength\tabcolsep{3.5pt}
  \begin{tabular}{lccc}
    \toprule
    Psuedo-coord  & \# latents & FID-CLIP($\downarrow$) & FID($\downarrow$) \\
    \midrule
    grid & 1024 & 7.32 & 9.74 \\
    random & 1024 & 11.66 & 17.99 \\ 
    KMeans++ & 1024 & 10.42 & 15.56 \\
    random & 2048 & 8.95 & 11.86 \\ 
    \arrayrulecolor{black}\bottomrule
  \end{tabular}
  }
  \caption{Performance of {\model} on protein folding.}
  \label{tab:spatial_aware_latent}
\end{table}

In image domain, we define pseudo coordinates to lie on a 2D grid, which results in pixels grouping as patches of same size. Tab.~\ref{tab:spatial_aware_latent} lists the results of an ablation study on LSUN-church-256 to compare different pseudo-coordinates. We trained all models for 200K steps with batch size 128 and report results in the table below. INRFlow achieves the best performance when using the default grid pseudo coordinates. When using randomly sampled pseudo coordinates we observe a drop in performance. 

We attribute this to the fact that when pseudo-coordinates are randomly sampled, each spatial-aware latent effectively does a different amount of work (since pixels only cross-attend to the nearest pseudo-coordinate). This unbalanced load across latents makes encoding less efficient. There are a few different ways to deal with this without necessarily relying on a grid, one is to cluster similar pseudo-coordinates to provide an equidistant distribution in 2D space (\ie KMeans++ initialization), another one is to increase the number of spatial-aware latents so that each latent has to do less work. We empirically see that both of this options are effective. Ultimately, having pseudo-coordinates lie on a grid strikes a good balance of efficiency and effectiveness.


\section{Architecture Ablation}

 \begin{table}[h]
  \centering
  \scalebox{0.84}{
  \setlength\tabcolsep{3.5pt}
  \begin{tabular}{lccc}
    \toprule
    Model  & FID$(\downarrow)$ & Precision$(\uparrow$) & Recall($\uparrow$) \\
    \midrule
    PerceiverIO & 65.09  & 0.38 & 0.01 \\
    \midrule
    {\model} (ours) & \textbf{7.03}  & \textbf{0.69} & \textbf{0.34} \\
    \arrayrulecolor{black}\bottomrule
  \end{tabular}
  }
  \caption{Benchmarking vanilla PerceiverIO and {\model} with spatially aware latents on  LSUN-Church-256 \citep{yu2015lsun}.}
  \label{tab:FID-latentvsambient}
\end{table}

We also provide an architecture ablation in Tab. \ref{tab:FID-latentvsambient} showcasing different design decisions. We compare two variants of Transformer-based architectures {\model}: a vanilla PerceiverIO that directly operates on ambient space, but without using spatial aware latents and {\model}. As it can be seen, the spatially aware latents introduced in {\model} greatly improve performance across all metrics in the image domain, justifying our design decisions. We note that we did not observe the same large benefits for 3D point clouds, which we hypothesize can be due to their irregular structure. 


\section{Unconditional 3D Point Cloud generation}
\label{sect:shapenet_exp}

\begin{table}[h]
  \centering
  \scalebox{0.7}{
  \setlength\tabcolsep{4pt}
  \begin{tabular}{lcccccccc}
    \toprule
    Model & Layers & Hidden size & \#Latents & Heads & Decoder layers & \#Params \\
    \midrule
    {\model}-B & 9 & 512 & 1024 & 4 & 1 & 108M \\
    {\model}-L & 12 & 512 & 1024 & 4 & 1 & 204M \\
    {\model}-XL & 28 & 1152 & 1024 & 16 & 1 & 866M \\
    \bottomrule
  \end{tabular}
  }
  \caption{Detailed configurations of {\model} for point cloud generation.}
  \label{tab:shapenet_model_config}
\end{table}

\begin{table*}[!t]                                
\centering  \small                            
\setlength{\tabcolsep}{6pt}                     
\begin{tabular}{cccccccc}                       
                                                
\toprule                                        
                                                &                                                &\multicolumn{2}{c}{MMD$\downarrow$}&\multicolumn{2}{c}{COV$\uparrow$ (\%)}&\multicolumn{2}{c}{1-NNA$\downarrow$ (\%)}              \\  \cmidrule(lr){3-4} \cmidrule(lr){5-6} \cmidrule(lr){7-8}
Category                                        &Model                                           &CD                  &EMD                 &CD                  &EMD                 &CD                  &EMD           \\  \midrule

                                                &ShapeGF~\citep{cai2020eccv}                     &0.3130     &0.6365     & \textbf{45.19}      &40.25               &81.23               &80.86             \\             
                                                &SP-GAN~\citep{li2021spgan}                       &0.4035              &0.7658              &26.42               &24.44               &94.69               &93.95             \\     
Airplane                                               &GCA~\citep{zhang2021learning}  &0.3586 &0.7651 &38.02  &36.30   &88.15  &85.93  \\ 
                                                &LION \citep{lion} (110M)                                          &0.3564              &0.5935              &42.96               &\textbf{47.90}      &76.30     &67.04            \\
                                                & \textbf{\model}-B (ours) (108M) & \textbf{0.2861} & 0.5156 & 43.38 & 47.54 & 75.55 & 64.95 \\
                                                & \textbf{\model}-L (ours)   & 0.2880 & \textbf{0.5052} & 44.44 & 47.16 & \textbf{62.20} & \textbf{62.96} \\

\midrule                                          
                                                &ShapeGF~\citep{cai2020eccv}                     &3.7243     &2.3944              &48.34      &44.26               &58.01               &61.25             \\              
                                                &SP-GAN~\citep{li2021spgan}                       &4.2084              &2.6202              &40.03               &32.93               &72.58               &83.69             \\           
Chair                                             &GCA~\citep{zhang2021learning}  &4.4035 &2.5820  &45.92  &47.89  &64.27  &64.50 \\ 
                                                &LION \citep{lion} (110M)                                         &3.8458              &2.3086     &46.37               &50.15     &56.50      &53.85            \\ 
                                                & \textbf{\model}-B (ours) (108M)  & 3.6310 & \textbf{2.1725} & 46.67 & \textbf{53.31} & 55.43 & \textbf{51.13} \\
                                                & \textbf{\model}-L (ours)  & \textbf{3.5145} & 2.1860 & \textbf{49.39} & 49.84 & \textbf{50.52} & 51.66 \\

\midrule                                          
                                                &ShapeGF~\citep{cai2020eccv}                     &1.0200     &0.8239              &\textbf{44.03}      &47.16               &61.79               &57.24             \\        
                                                &SP-GAN~\citep{li2021spgan}                       &1.1676              &1.0211              &34.94               &31.82               &87.36               &85.94             \\     
Car                                               &GCA~\citep{zhang2021learning} &1.0744 &0.8666 &42.05  &48.58  &70.45  &64.20 \\ 
                                                &LION \citep{lion}    (110M)                                   &1.0635              &0.8075     &42.90               &\textbf{50.85}      &59.52      &\textbf{49.29}            \\  
                                                & \textbf{\model}-B (ours) (108M)   & 0.9923 & \textbf{0.7692} & 43.46 & 47.44 & 60.36 & 53.27 \\
                                                & \textbf{\model}-L (ours) & \textbf{0.9660} & 0.7846 & \textbf{44.03} & 48.86 & \textbf{53.83} & 54.55 \\

\midrule

& LION \citep{lion} (110M)  &  3.4336     &      2.0953      &    48.00    &     52.20   &      58.25      &    57.75   \\ 

All (55 cat) & \textbf{\model}-B (ours) (108M) &  3.2586   &      2.1328     &   49.00  &     50.40   &  54.65  &   55.70   \\ 

& \textbf{\model}-L (ours)  &  \textbf{3.1775}    &      \textbf{1.9794 }     &    \textbf{49.80}  &     \textbf{52.39}   &      \textbf{51.80}      &    \textbf{53.90}   \\

\bottomrule

\end{tabular}                                   
\caption{Generation performance metrics on Airplane, Chair, Car and all 55 categories jointly. All models were trained on the ShapeNet dataset from PointFlow \citep{yang2019pointflow}. Both the training and testing data are normalized individually into range [-1, 1]. }
\label{tab:gen_v1_individual_norm_full}                 
\end{table*}

For completeness we also tackle unconditional 3D point cloud generation on ShapeNet \citep{shapenet}. Note that our model does not require training separate VAEs for point clouds, tuning their corresponding hyper-parameters or designing domain specific networks. We simply adapt our architecture for the change in dimensionality of coordinate-value pairs (\eg $f: \mathbb{R}^2 \rightarrow \mathbb{R}^3$ for images to $f: \mathbb{R}^3 \rightarrow \mathbb{R}^3$ for 3D point clouds.). Note that for 3D point clouds, the coordinates and values are equivalent. In this setting, we compare baselines including LION \citep{lion} which is a recent state-of-the-art approach that models 3D point clouds using a latent diffusion type of approach. Following \cite{lion} we report MMD, COV and 1-NNA as metrics. To have a straightforward comparison with baselines, we train {\model}-B with to approximately match the number of parameters as LION~\citep{lion} (110M for LION vs 108M for {\model}, see Tab. \ref{tab:shapenet_model_config}) on the same datasets (using per sample normalization as in Tab. 17 in \citet{lion}). On ShapeNet, {\model} models are trained for 800K iterations with a batch size of 16.

We show results for category specific models and for an unconditional model jointly trained on 55 ShapeNet categories in Tab.~\ref{tab:gen_v1_individual_norm_full}. {\model}-B obtains strong generation results on ShapeNet despite being a domain agnostic approach and outperforms LION in most datasets and metrics. Note that {\model}-B has comparable number of parameters and the same inference settings than LION so this is fair comparison. Finally, we also report results for a larger model {\model}-L (with $\times 2$ the parameter count as LION) to investigate how {\model} improves as with increasing model size. We observe that with increasing model size, {\model} typically achieves better performance than the base version. This further demonstrates scalability of our model on ambient space of different data domains. More point cloud samples can be found in Appendix~\ref{app:shapenet_samples}.


\section{Quantitative Results on Protein Folding}

 \begin{table}[h]
  \centering
  \scalebox{0.84}{
  \setlength\tabcolsep{3.5pt}
  \begin{tabular}{lcc}
    \toprule
    Model  & $C_\alpha$-LDDT($\uparrow$) & TM-Score($\uparrow$) \\
    \midrule
    Boltz1~\cite{wohlwend2024boltz} & 0.923 & 0.812 \\
    {\model} & 0.722 & 0.664 \\
    \arrayrulecolor{black}\bottomrule
  \end{tabular}
  }
  \caption{Performance of {\model} on protein folding.}
  \label{tab:folding}
\end{table}

This section includes the quantitative evaluation of the protein folding task. In particular, we randomly selected 512 proteins from the AFDB-SwissProt dataset~\cite{afdb} and use them as a test set. We compare {\model} with an open-source replication of AlphaFold3~\cite{alphafold3} (ie. Boltz-1~\cite{wohlwend2024boltz}), which is the SOTA approach for protein folding. Noted that AlphaFold3 is extremely domain-specific, using complex and curated architectural designs for protein folding. For example, it relies on multi-sequence alignment and template search on existing proteins. It also designs a triangle self-attention update for atomic coordinates. Whereas INRFlow makes no assumptions about data domain and effectively models different tasks under an equivalent architecture and training objective. We report $C_{\alpha}$-LDDT and TM-score which are commonly used metrics to evaluate how predicted protein structures align with ground truth (Tab.~\ref{tab:folding}). Results indicate that INRFlow, which uses a domain-agnostic architecture performs decently well on protein folding even when compared to SOTA models that require intricate domain-expertise embedded in the architecture. Note that we have not optimized INRFlow for hyper-parameters in the protein folding experiment.


\section{Implementation of Resolution Agnostic Generation}

\begin{figure}[htb!]
\centering
\includegraphics[width=0.45\textwidth]{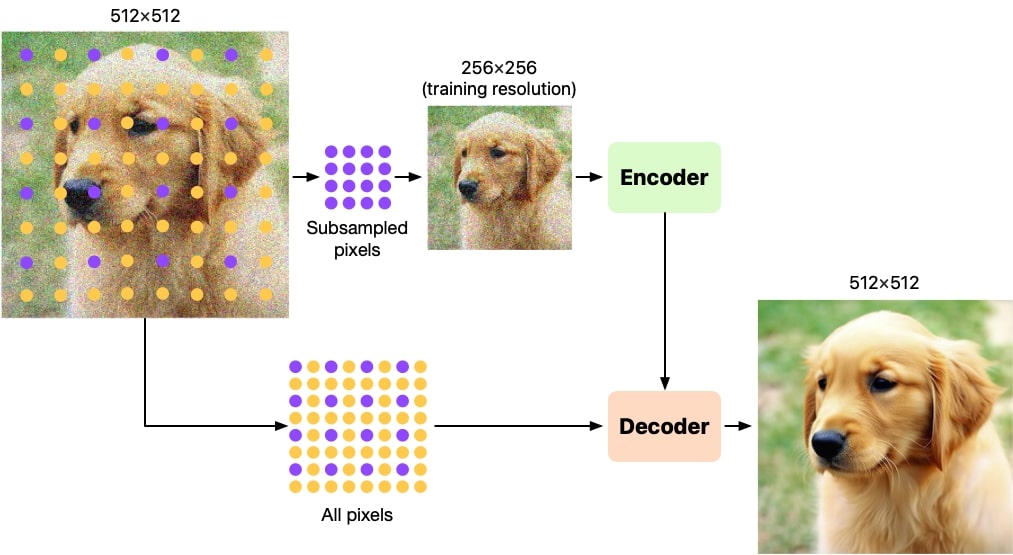}
\caption{Illustration of resolution agnostic sampling. When generating higher-resolution images in inference (e.g., 512$\times$512), 256$\times$256 coordinate-value pairs (consistent to setting in training)  are selected through grid subsampling and are fed to the encoder. The decoder takes in the full 512$\times$512 coordinate-value pairs to predict velocities of all the pixel values. The model repeats the process in inference to generate a 512$\times$512 image.}
\label{fig:super_res_illustrate}
\end{figure}

One can interpret the spatial-aware latents computed from INRFlow's encoder as the ``latent codes that are transformed into network parameters''. For INRFlow these latents codes are used to compute $v,k$ for the cross-attention block in the decoder, which takes in queries $q$ (\ie coordinate-value pairs) at any resolution. The only thing that we need to do in order to obtain consistent outputs across different query resolutions is to keep the resolution of the encoder fixed, while the decoder can be queried at a different resolution. This is trivially achieved by employing a simple sub-sampling operation (\ie grid sub-sampling) and keeping the sub-sampling operation fixed during inferenc (Fig.~\ref{fig:super_res_illustrate}). This simple technique allows us to change the resolution at inference time without any other additional tricks regarding noise alignment at different resolutions and produces crisp and consistent examples at higher resolutions that the one used in training (more results see Fig.~\ref{fig:resolution_agnostic} and \ref{fig:image_upsample_appendix}).


\section{Additional ImageNet Samples}
\label{app:imagenet_samples}

\begin{figure*}[h]
\centering

\setlength{\tabcolsep}{6pt}
\resizebox{0.85\textwidth}{!}{
    \begin{tabular}{cc}
    \centering
    \includegraphics[width=0.49\textwidth]{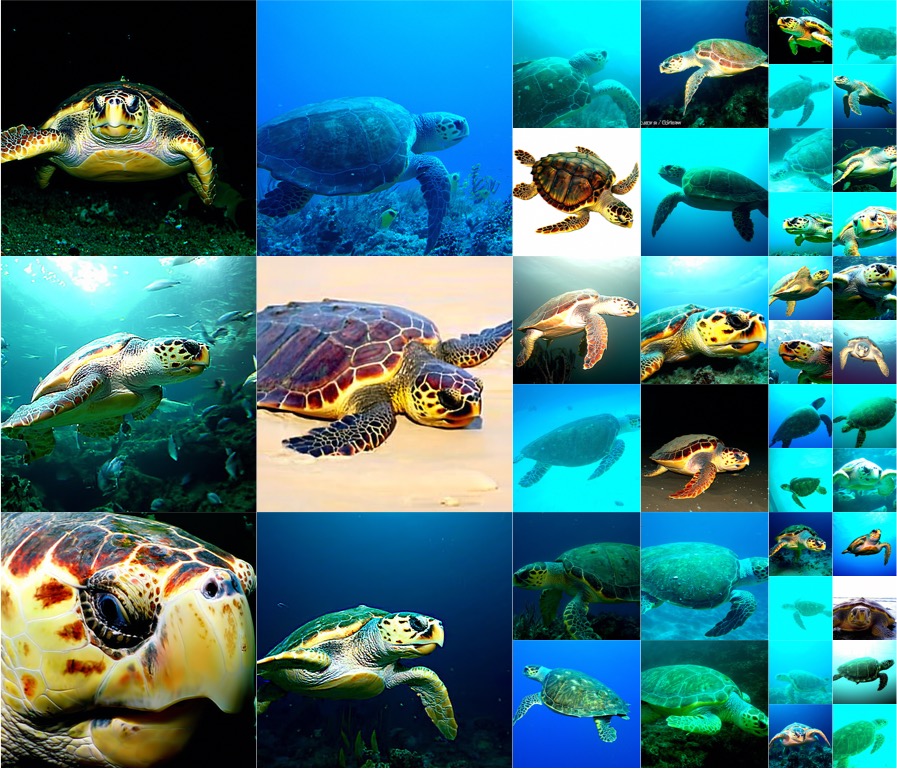} & \includegraphics[width=0.49\textwidth]{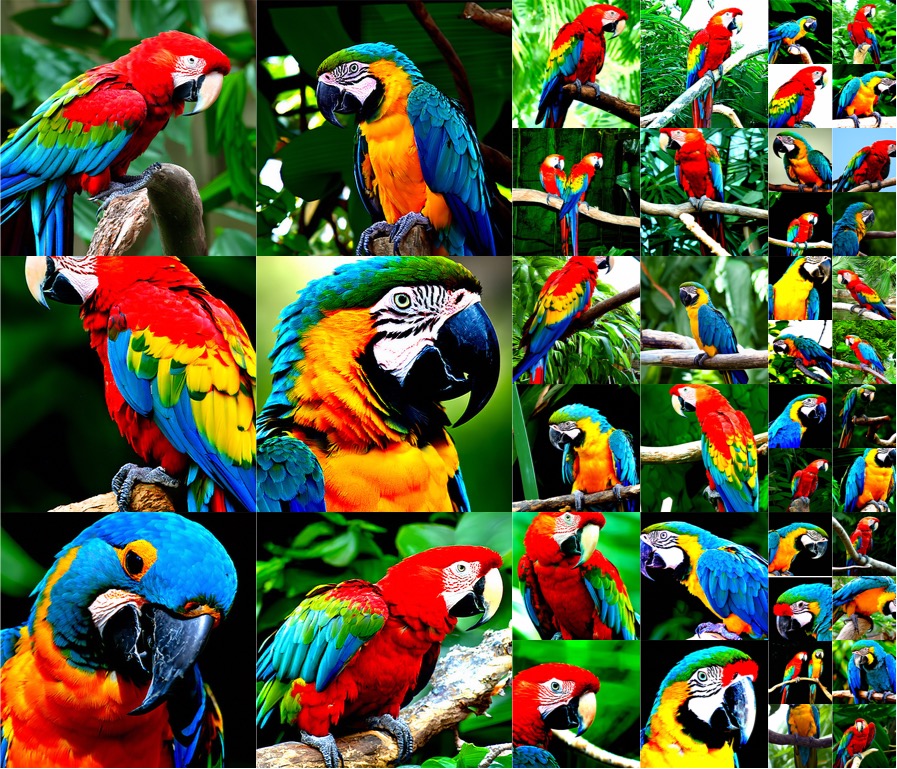} \\
    \includegraphics[width=0.49\textwidth]{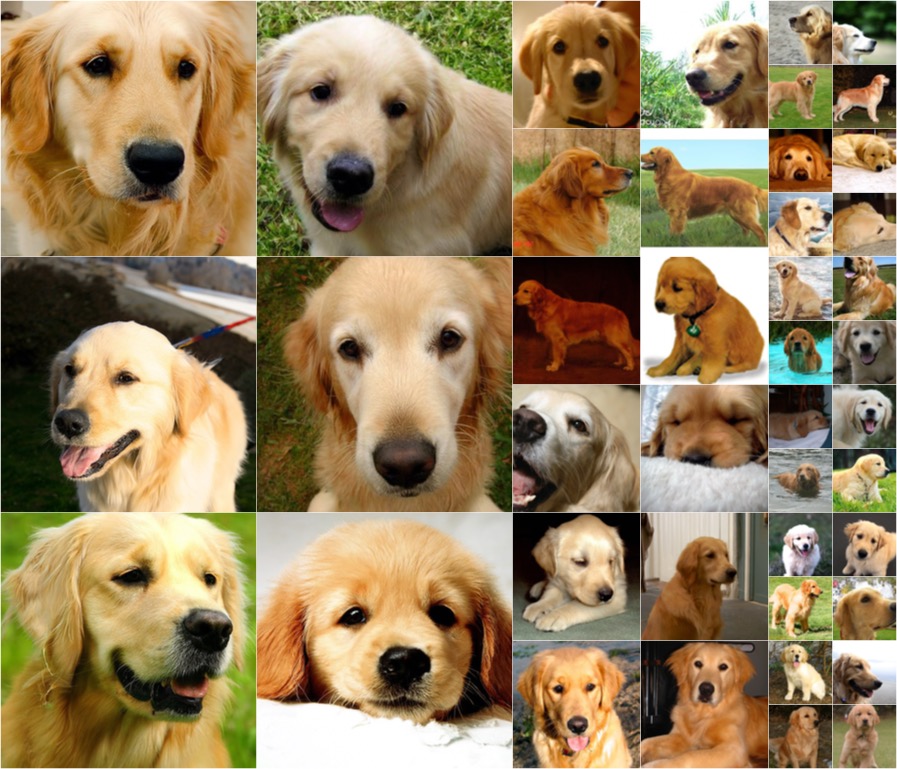} & 
    \includegraphics[width=0.49\textwidth]{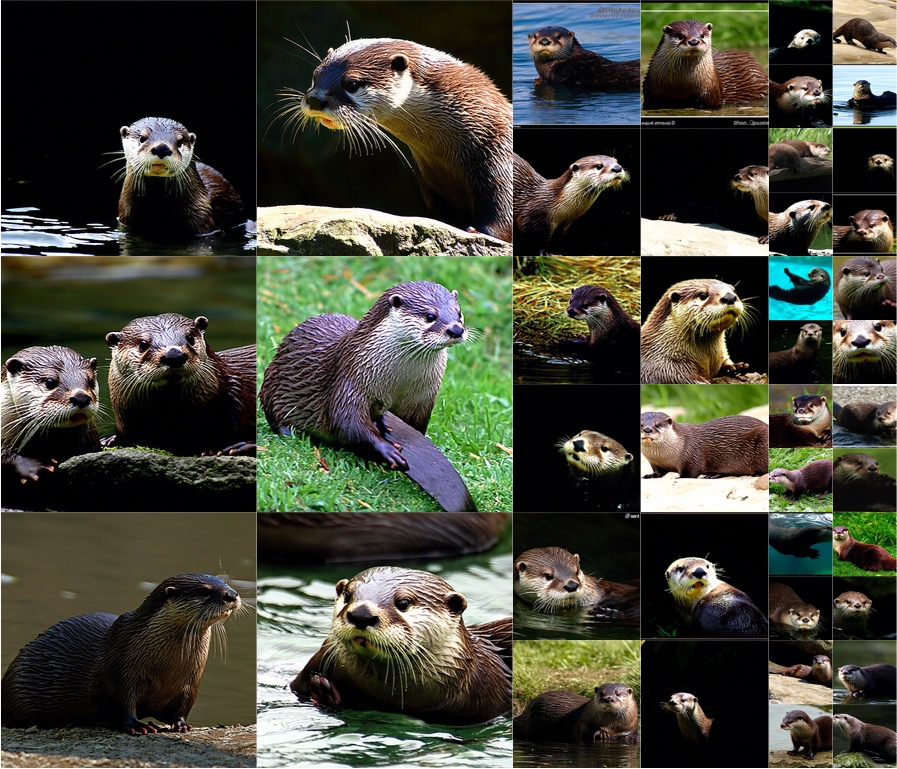} \\
    \includegraphics[width=0.49\textwidth]{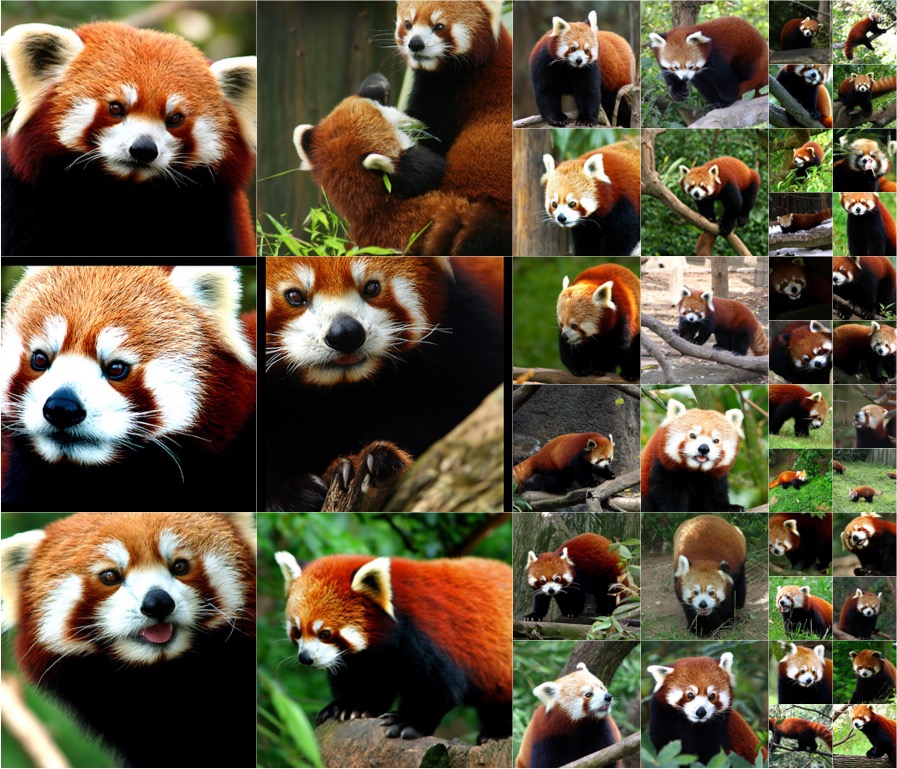} &
    \includegraphics[width=0.49\textwidth]{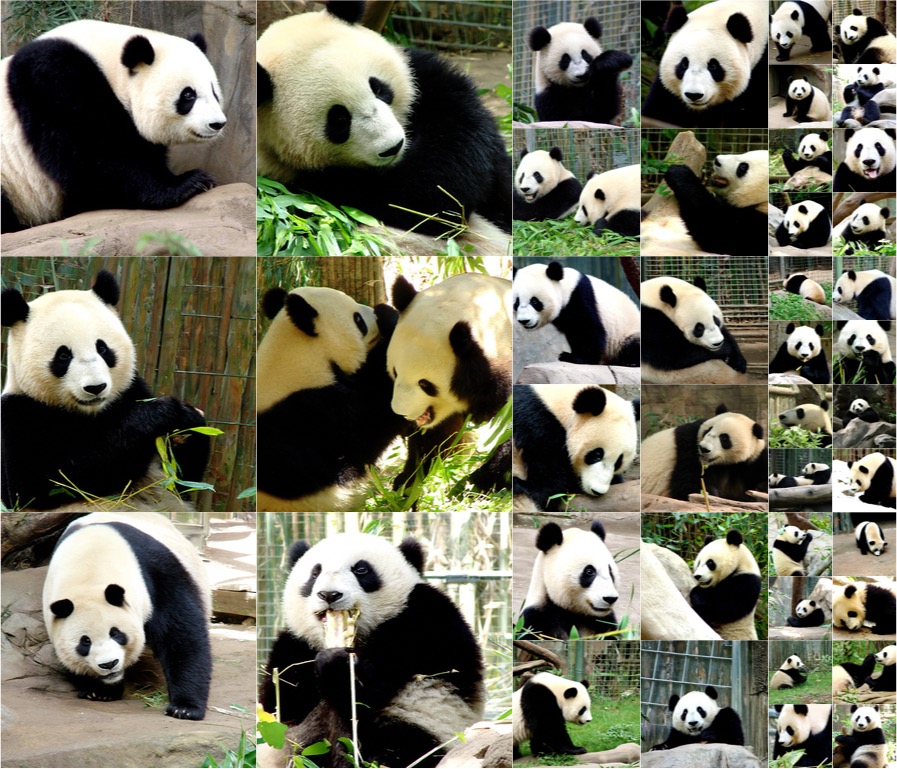} \\
    \end{tabular}
    }
    \caption{Uncurated samples of class labels: loggerhead turtle (33), macaw (88), golden retriever (207), otter (360) and red panda (387), and panda (388) from {\model} trained on ImageNet-256.}
    \label{fig:imagenet_samples1}
\end{figure*}

\begin{figure*}[h]
\centering
\setlength{\tabcolsep}{6pt}
\resizebox{0.85\textwidth}{!}{
    \begin{tabular}{cc}
    \centering
    \includegraphics[width=0.49\textwidth]{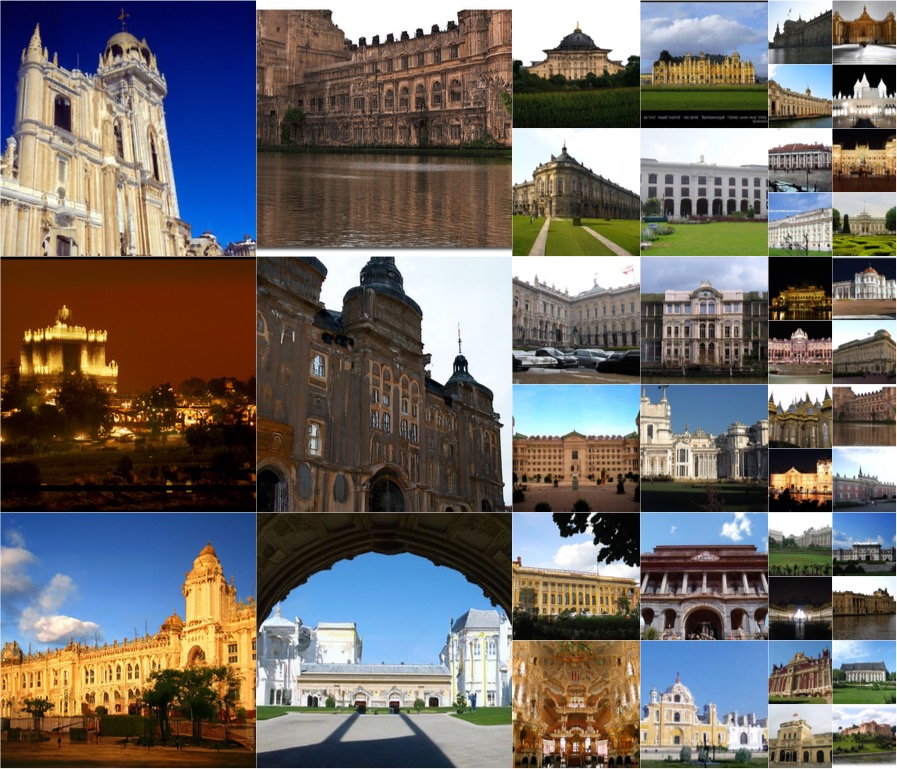} & \includegraphics[width=0.49\textwidth]{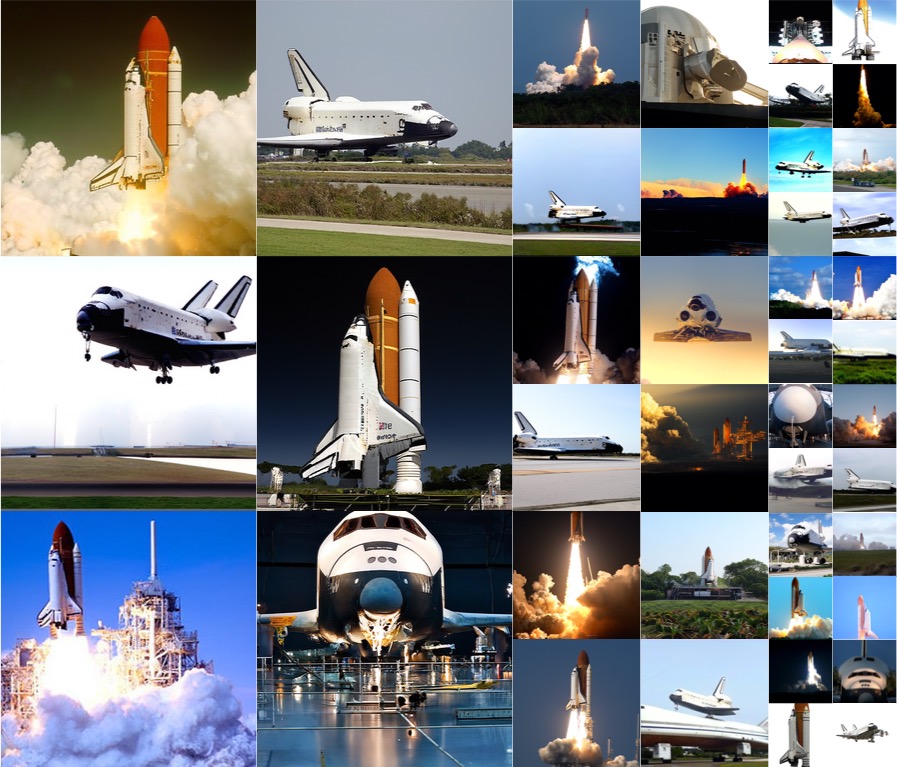} \\
    \includegraphics[width=0.49\textwidth]{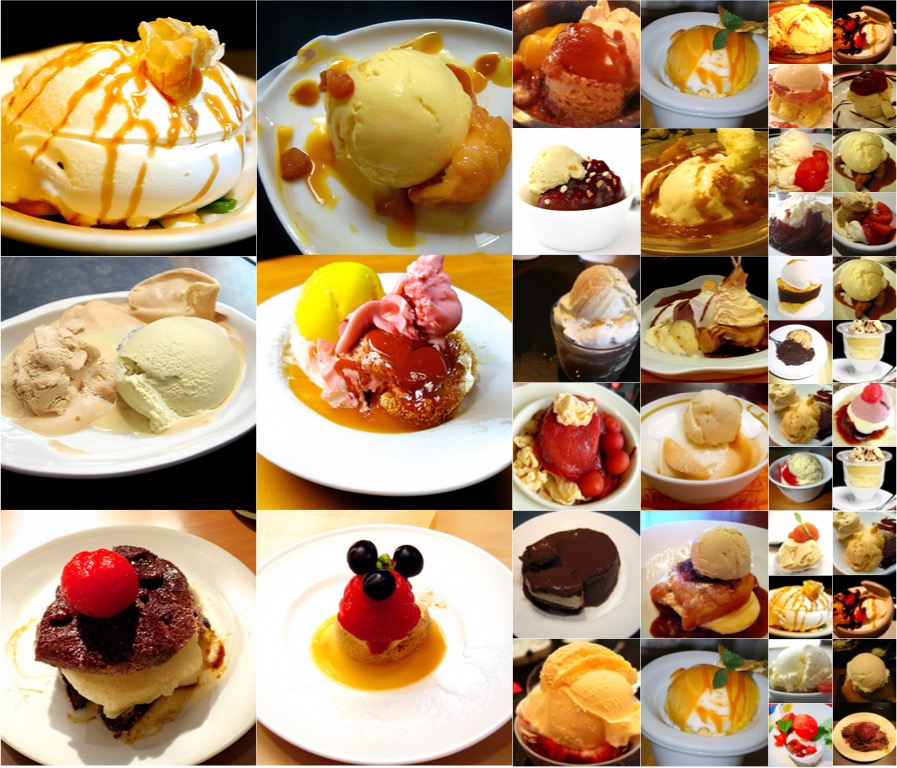} & \includegraphics[width=0.49\textwidth]{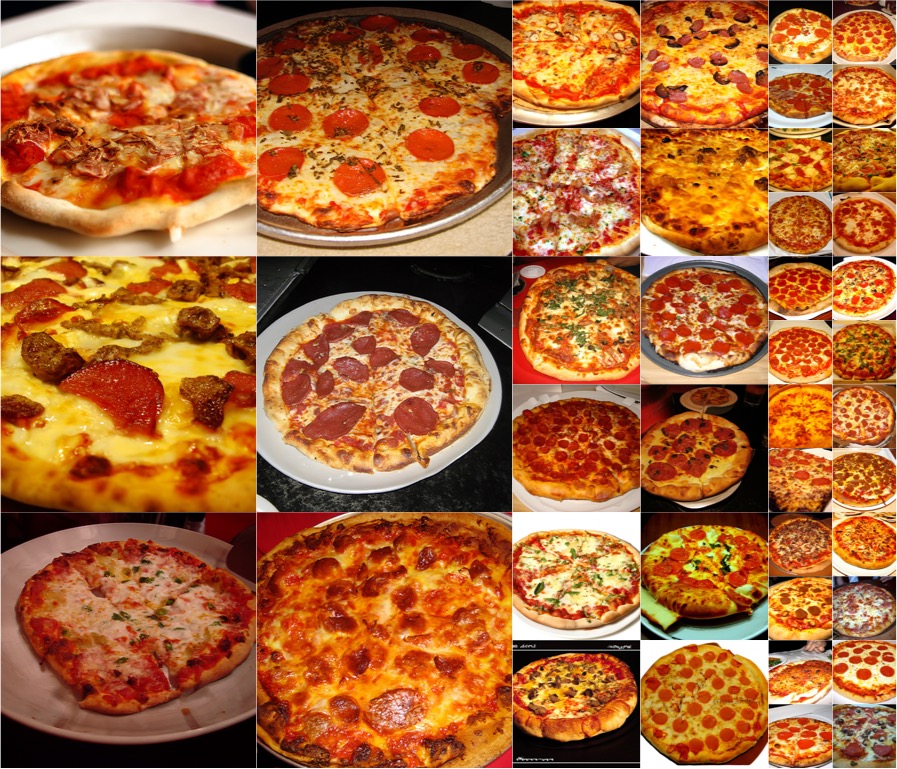} \\
    \includegraphics[width=0.49\textwidth]{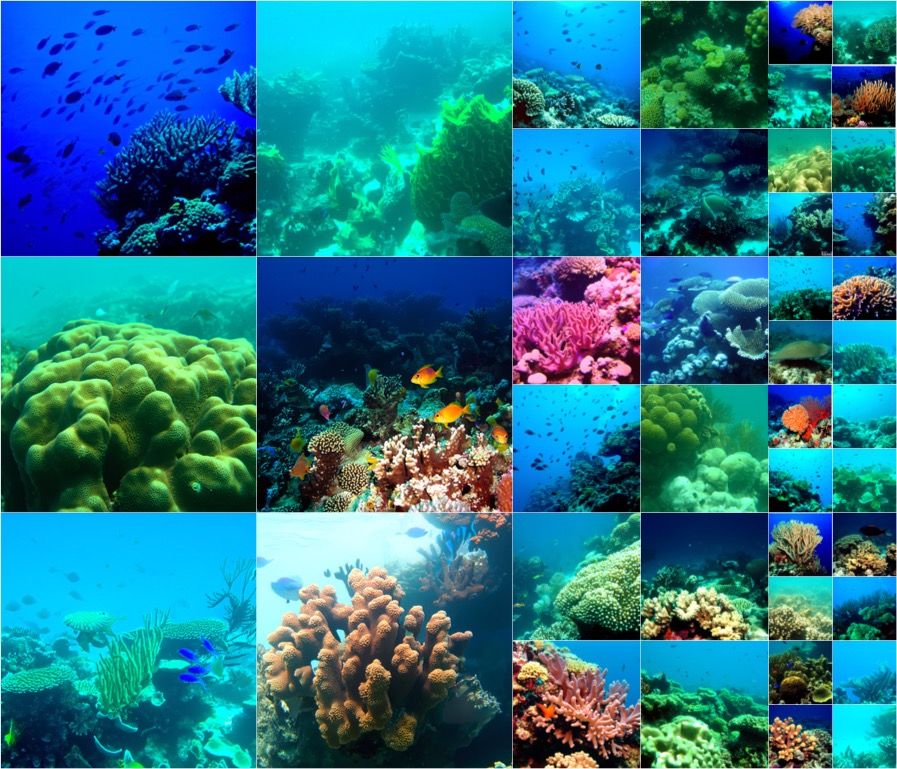} & \includegraphics[width=0.49\textwidth]{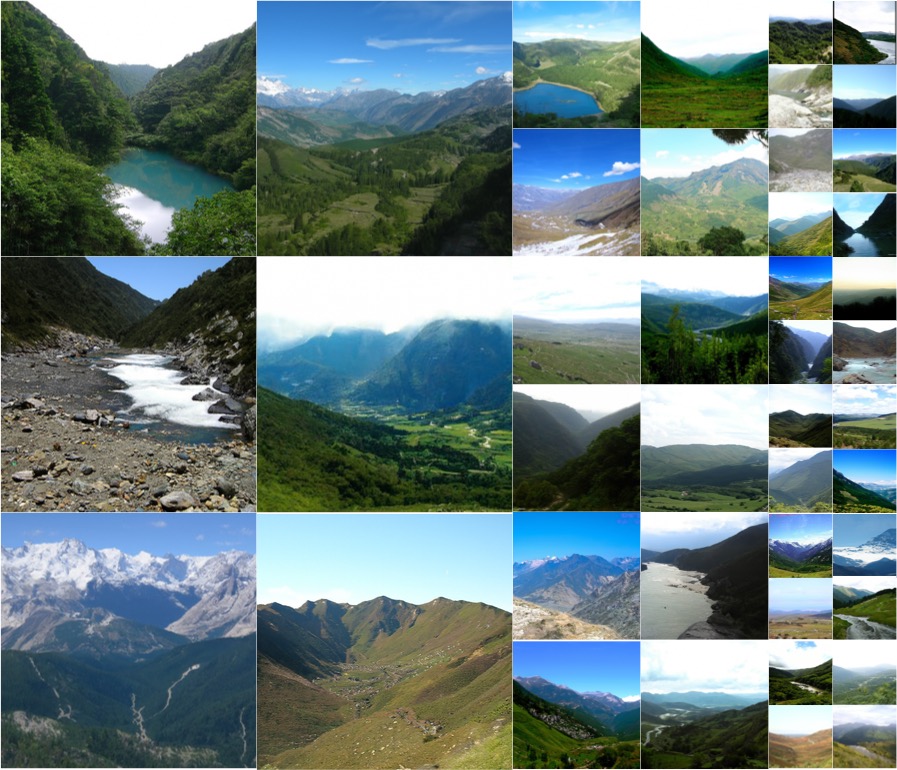} \\
    \end{tabular}
    }
    \caption{Uncurated samples of class labels: palace (698), space shuttle (812), ice cream (928), pizza (963), coral reef (973), and valley (979) from {\model} trained on ImageNet-256.}
    \label{fig:imagenet_samples2}
\end{figure*}

We show uncurated samples of different classes from {\model}-XL trained on ImageNet-256 in Fig. \ref{fig:imagenet_samples1} and Fig. \ref{fig:imagenet_samples2}. Guidance scales in CFG are set as $4.0$ for loggerhead turtle, macaw, otter, coral reef and $2.0$ otherwise. 

\section{Additional ShapeNet Samples}
\label{app:shapenet_samples}
\begin{figure*}
    \centering
    \includegraphics[width=0.9\textwidth]{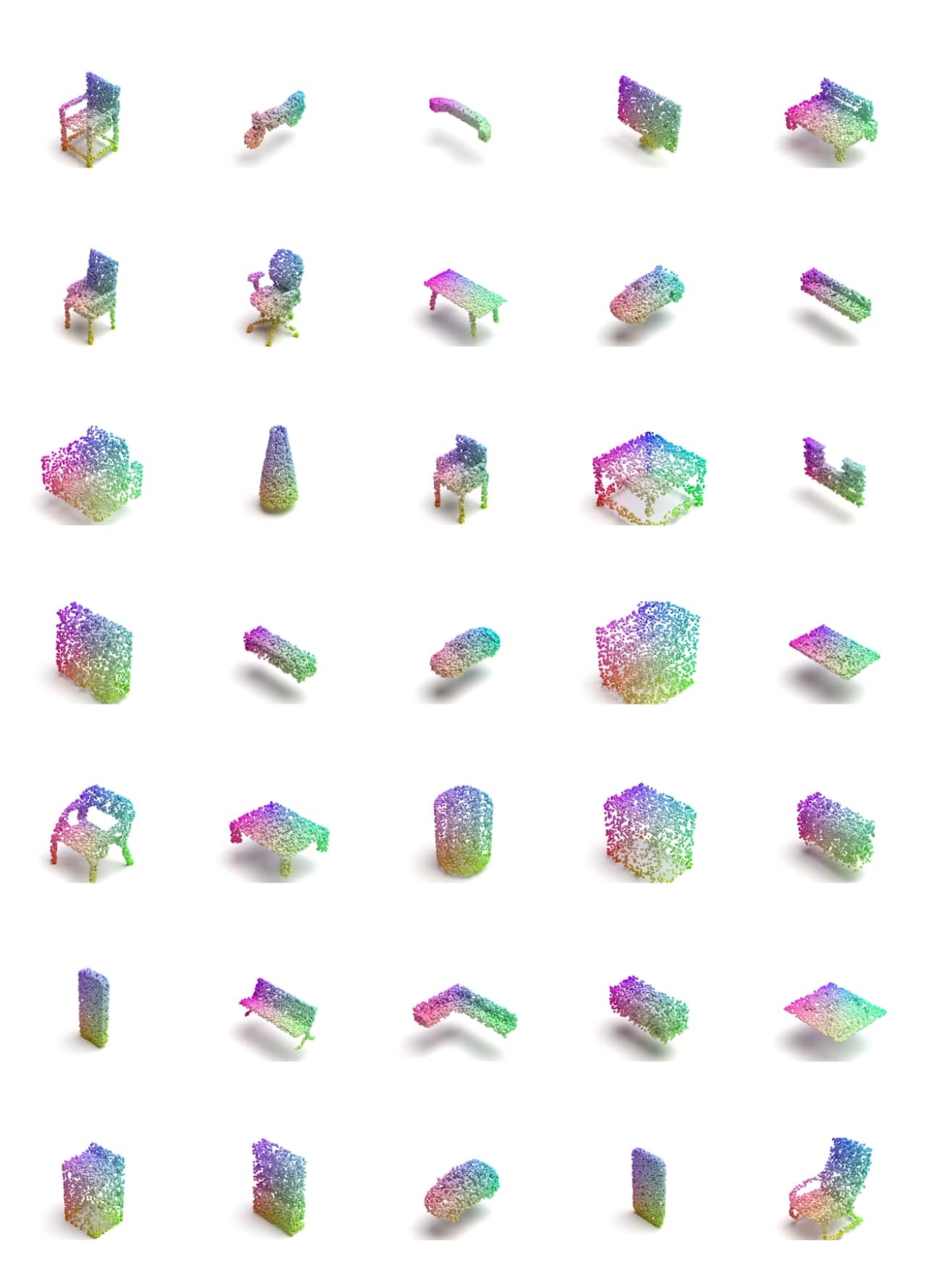}
    \caption{Additional uncurated ShapeNet generations using $2048$ points from the unconditional model jointly trained on 55 categories}
    \label{fig:app_shapenet_1}
\end{figure*}

We show uncurated samples from {\model}-L trained jointly on 55 ShapeNet categories in Fig. \ref{fig:app_shapenet_1}.

\section{Additional Objaverse Samples}
\begin{figure*}[!h]
    \centering
    \includegraphics[width=\linewidth]{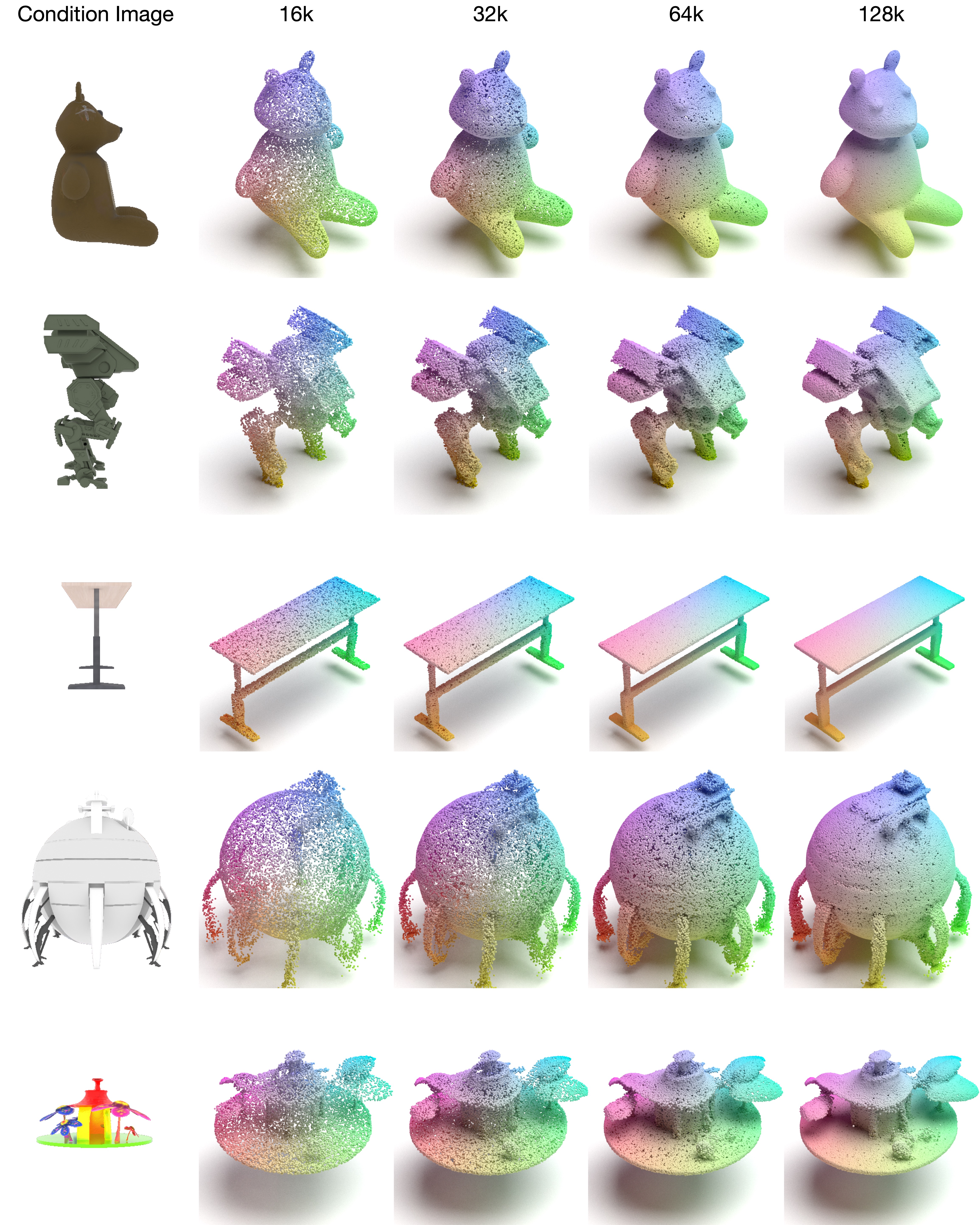}
    \caption{Image-conditioned point clouds with 16k, 32k, 64k, and 128k points generated from an {\model} trained on Objaverse (training with 16k points, CFG scale 5.0). }
    \label{fig:app_objaverse}
\end{figure*}

We show additional Objaverse samples from {\model} in Fig. \ref{fig:app_objaverse}

\section{Additional SwissProt  Samples}
\begin{figure*}[!h]
    \centering
    \includegraphics[width=0.8\linewidth]{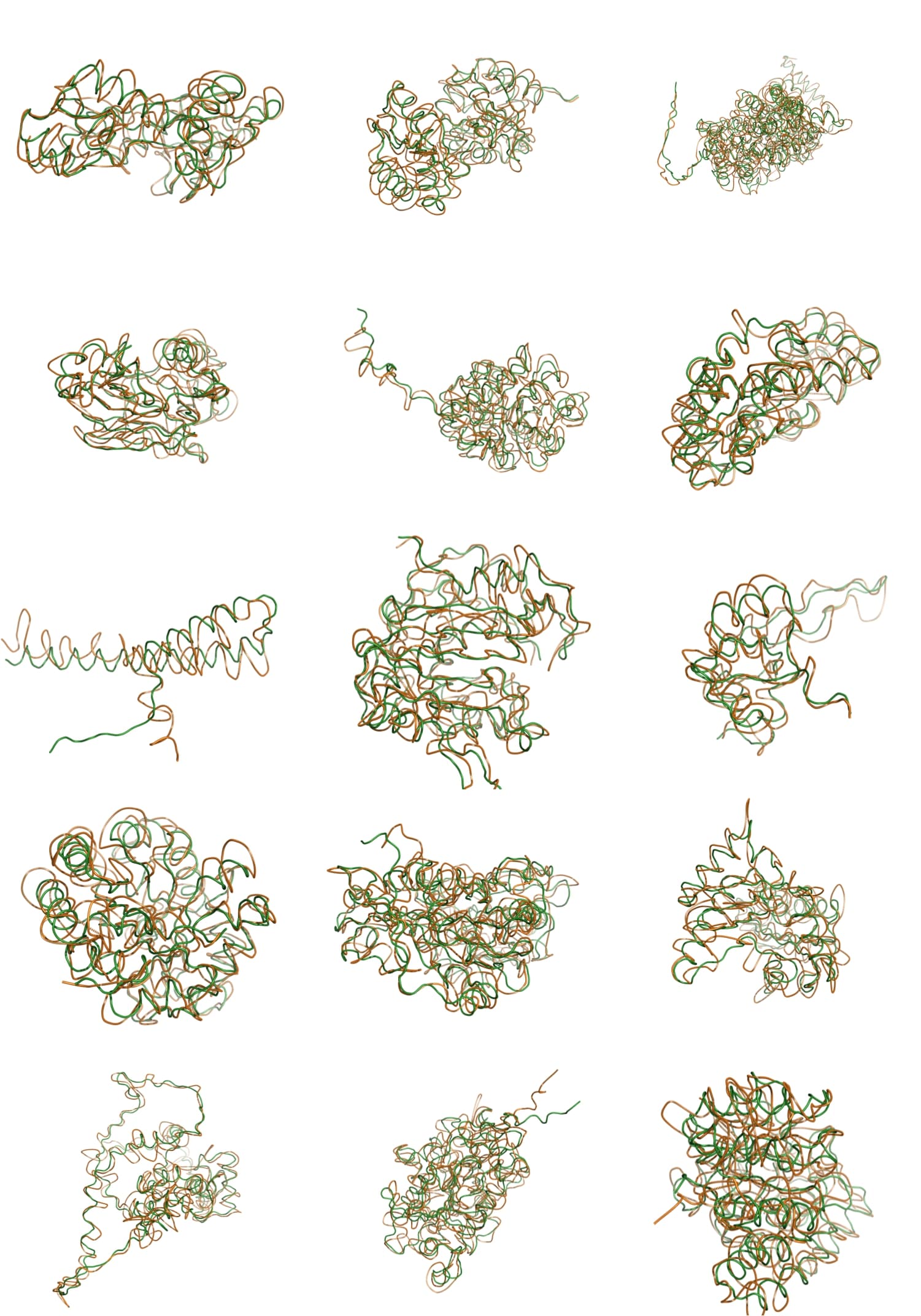}
    \caption{Additional examples of protein structures predicted by {\model} on SwissProt \citep{boeckmann2003swiss}}
    \label{fig:app_folding}
\end{figure*}

We show additional SwissProt samples from {\model} in Fig. \ref{fig:app_folding}

\section{Additional Resolution Agnostic Image Samples}

\begin{figure*}
    \centering
    \includegraphics[width=0.85\textwidth]{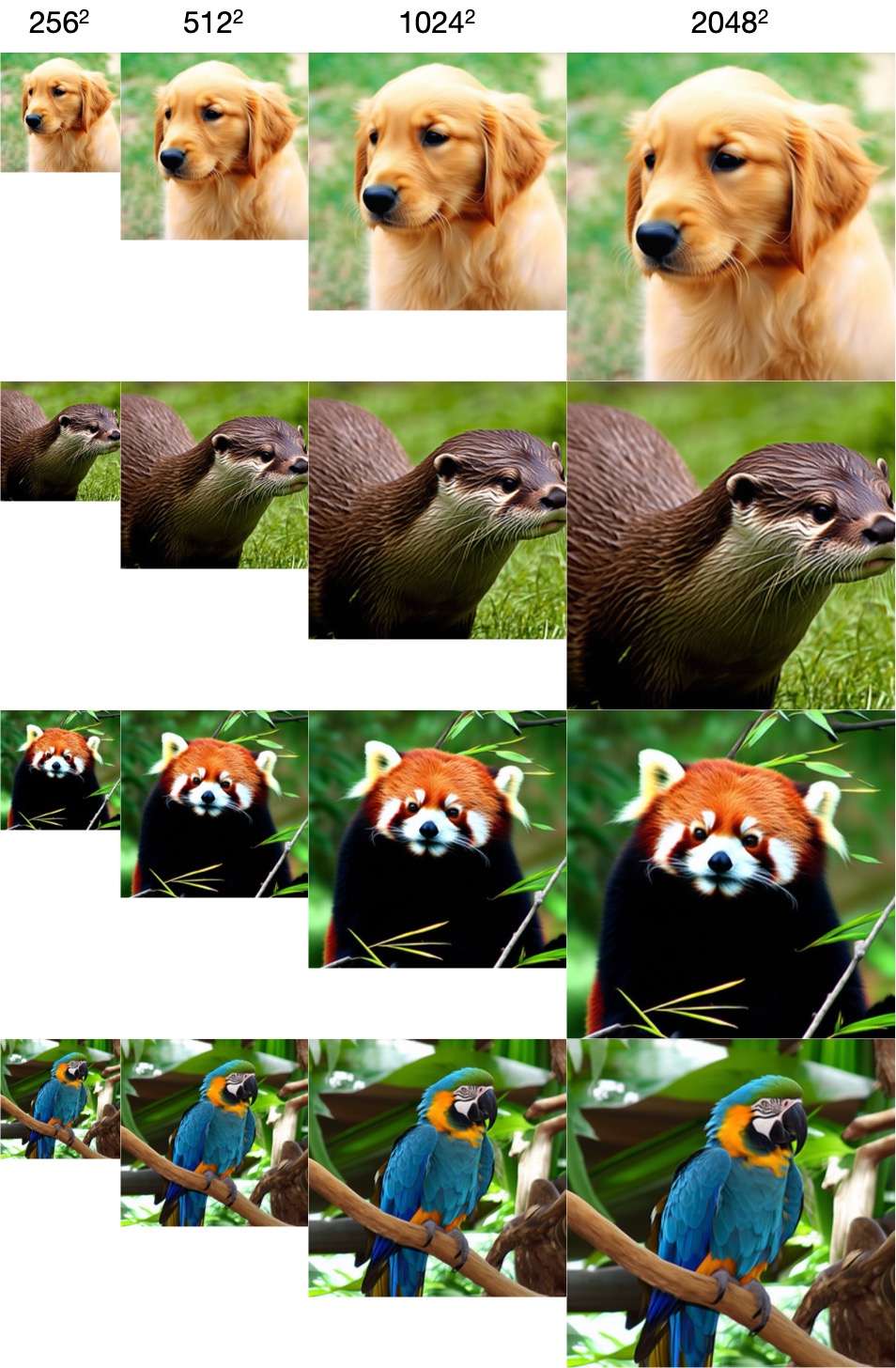}
    \caption{\textcolor{black}{Images generated at 256, 512, 1024, and 2048 resolutions from an {\model} trained on ImageNet-256.} }
    \label{fig:image_upsample_appendix}
\end{figure*}

\textcolor{black}{
We show additional samples generated at different resolutions from {\model} trained on ImageNet-256 in Fig.~\ref{fig:image_upsample_appendix}.
}

\end{document}